\documentclass[runningheads]{llncs}
\usepackage[T1]{fontenc}
\usepackage{times}
\usepackage{hyperref}
\usepackage{multicol}
\usepackage{multirow}
\usepackage{colortbl}
\usepackage{wrapfig}
\usepackage{amsmath}
\usepackage{subcaption}
\usepackage{booktabs} 
\usepackage{xcolor}
\usepackage{floatrow}
\usepackage{graphicx}
\usepackage{color}
\usepackage{amssymb} 
\usepackage{pifont}  
\usepackage[flushleft]{threeparttable}
\usepackage[ruled,linesnumbered,vlined]{algorithm2e}

\newcommand{\firsttone}[1]{\colorbox{red!13}{#1}}
\newcommand{\secondtone}[1]{\colorbox{green!13}{#1}}
\newcommand{\thirdtone}[1]{\colorbox{blue!13}{#1}}
\newfloatcommand{capbtabbox}{table}[][\FBwidth]
\newcommand{\cmark}{{\color{black}\ding{51}}}
\newcommand{\cgmark}{{\color{green}\ding{51}}}
\newcommand{\cbmark}{{\color{blue}\ding{51}}}
\newcommand{\cymark}{{\color{yellow}\ding{51}}}
\newcommand{\xmark}{{\color{gray}\ding{55}}}
\SetAlFnt{\small}
\SetAlCapFnt{\small}
\SetAlCapNameFnt{\small}
\newcommand{\var}{\texttt}
\let\oldnl\nl 
\newcommand{\nonl}{\renewcommand{\nl}{\let\nl\oldnl}} 

\begin{document}
%
\title{Unified-EGformer: Exposure Guided Lightweight Transformer for Mixed-Exposure Image Enhancement}
\titlerunning{Unified-EGformer}
%
\author{
    Eashan Adhikarla\inst{1}\orcidID{0009-0006-8417-0788}
    Kai Zhang\inst{1}\orcidID{0000-0002-6322-6096} \\
    Rosaura G. VidalMata\inst{2}\orcidID{0000-0002-4096-3747}
    Manjushree Aithal\inst{2}\orcidID{0000-0001-8953-5906}
    Nikhil Ambha Madhusudhana\inst{2}\orcidID{0009-0003-8272-6760} \\
    John Nicholson\inst{2}\orcidID{0009-0005-3193-9082}
    Lichao Sun \inst{1}\orcidID{0000-0003-1539-7939} \\
    Brian D. Davison\inst{1}\orcidID{0000-0002-9326-3648}
}

\authorrunning{Adhikarla et al.}
%
\institute{
    Lehigh University, Bethlehem PA 18015, USA 
    \email{\{eaa418,kaz321,lis221,bdd3\}@lehigh.edu}\\
    \and
    Lenovo Research, USA\\
    \email{\{rosaurav,maithal,amnikhil,jnichol\}@lenovo.com}
}

\maketitle 
\setcounter{footnote}{0}
%
\begin{abstract}
    Despite recent strides made by AI in image processing, the issue of mixed exposure, pivotal in many real-world scenarios like surveillance and photography, remains a challenge. Traditional image enhancement techniques and current transformer models are limited with primary focus on either overexposure or underexposure. To bridge this gap, we introduce the Unified-Exposure Guided Transformer (\textbf{Unified-EGformer}). Our proposed solution is built upon advanced transformer architectures, equipped with local pixel-level refinement and global refinement blocks for color correction and image-wide adjustments. We employ a guided attention mechanism to precisely identify exposure-compromised regions, ensuring its adaptability across various real-world conditions. U-EGformer, with a lightweight design featuring a memory footprint (peak memory) of only $\sim$\textbf{1134 MB} (0.1 Million parameters) and an inference time of 95 ms (over \textbf{9x} faster than typical existing implementations, is a viable choice for real-time applications such as surveillance and autonomous navigation. Additionally, our model is highly generalizable, requiring minimal fine-tuning to handle multiple tasks and datasets with a single architecture. 
\end{abstract}

\section{Introduction}
\label{sec:intro}
    \begin{figure}
    \centering
        \includegraphics[scale=0.225]{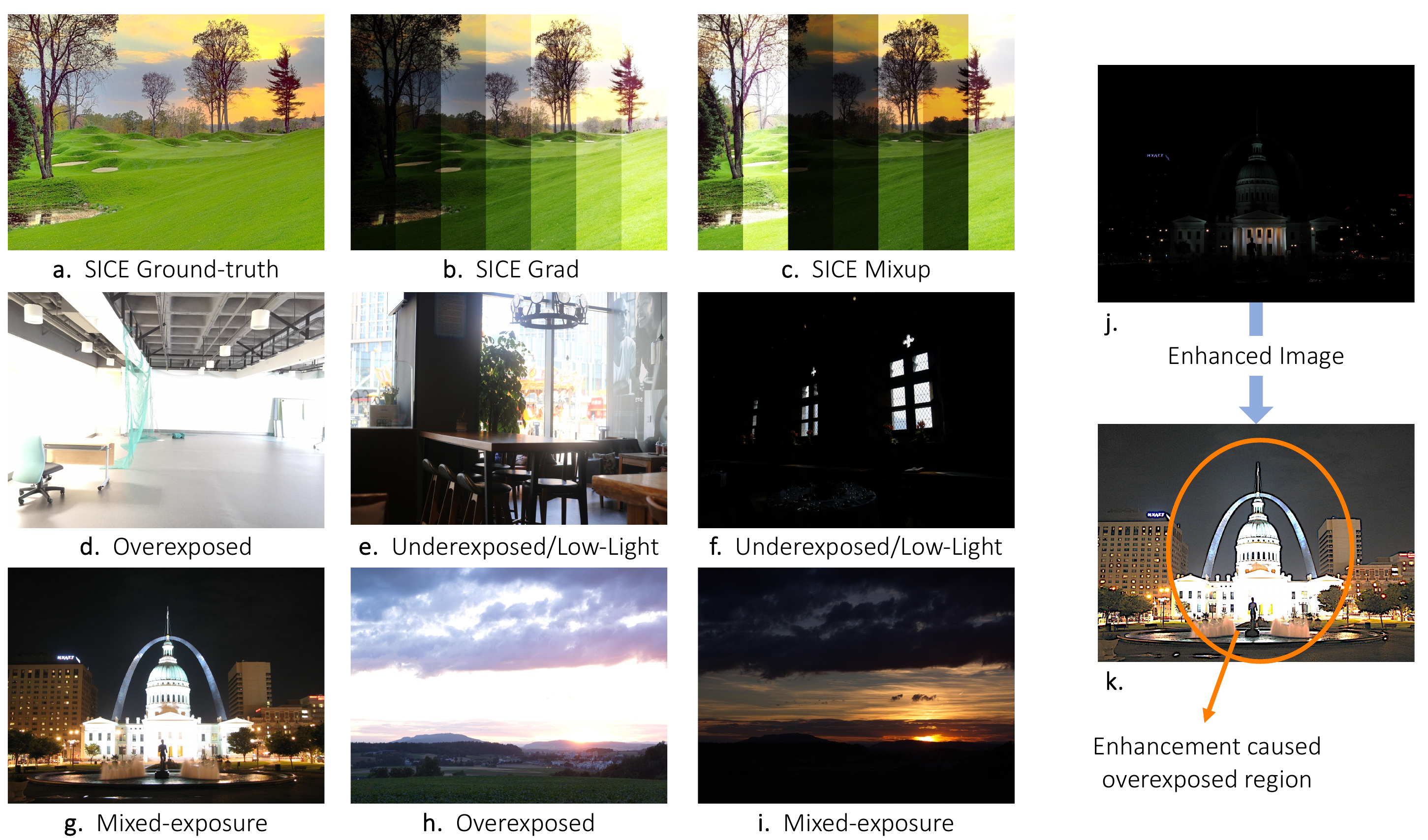}
        \caption{Sub-figures [a,b,c] show the handcrafted mixed exposure dataset by Zheng et al.~\cite{zheng2022low}; images [d-i] from Cai et al.~\cite{Cai2018deep} illustrate real-world scenarios of underexposure, overexposure, and mixed-exposures. Images [j,k] demonstrate the problem practically.}
        \label{fig:intro-to-problem}
    \end{figure}
    AI-driven image processing has significantly broadened the scope for enhancing visual media quality. A critical challenge within low-light image enhancement is addressing \textbf{\textit{mixed exposure}} in images as in Fig.~\ref{fig:intro-to-problem}~g., where a single frame contains both underlit\footnote{insufficient brightness in an image where details are lost due to lack of signal.} (below 5 lux, including underexposed) and overlit\footnote{excessive brightness in an image where details are lost due to signal clipping or saturation.} (overexposed) regions. This issue extends beyond academic interest and has significant real-world implications. For instance, in video calls (e.g., in cafeterias as in Fig.~\ref{fig:intro-to-problem}~e.) and live streaming, low-light enhancement is pivotal for clear visual communication. Other areas of application include autonomous driving, surveillance and security, photography, etc. Professional photographers often use high-end DSLR cameras and meticulously adjust settings such as aperture, ISO, and utilize specialized filters to mitigate exposure issues. However, such pre- and post-processing techniques are often not practical.

    Existing methods for correcting mixed exposure images have typically treated underexposure \cite{guo2020zero,moran2020deeplpf,Wang_2019_CVPR} and overexposure as separate challenges within the low-light image enhancement task. Although these methods (such as FECNet \cite{10.1145/3343031.3350855}, ELCNet+ERL \cite{huang2023learning}, IAT \cite{Cui_2022_BMVC}) have made progress, they commonly assume uniform scene illumination, leading to global adjustments that either brighten or darken the entire image. Such approaches falls short when dealing with images that have both overexposed and underexposed regions due to non-uniform lighting, resulting in suboptimal performance. For example, ZeroDCE \cite{guo2020zero} and RUAS \cite{ll_benchmark} can worsen overexposure in background regions while trying to enhance underexposed foreground subjects as shown in Fig.~\ref{fig:intro-to-problem}~k.

    We see this as a gap in the literature that has not been fully addressed, despite some efforts such as LCDNet \cite{wang2022local} and night enhancement approaches \cite{jin2022unsupervised,Cai2018deep,afifi2021learning}. The motivation for our work arises from the need to address these limitations with a solution that can handle mixed exposure scenarios effectively and is suitable for deployment on edge devices. An additional challenge when applying a low-light enhancement approach to a multi-exposed image is when a model trained solely on underexposed paired images inadvertently exacerbates overexposed regions, as seen in  Fig.~\ref{fig:intro-to-problem}~k, and vice versa. Consequently, the issue of mixed exposure emerges as a pivotal yet largely unexplored frontier. Notably, mobile phone cameras face this issue acutely, requiring lightweight, low-latency models that can operate within the device’s resource constraints while delivering high-quality image enhancement. Our goal is to develop a solution that not only addresses these issues but also offers faster inference and a lower memory footprint, making it ideal for practical applications on edge devices. 
    
    This paper develops an effective and computing-efficient approach to tackle mixed exposure in low-light image enhancement. Our major contributions are as follows:
    \begin{enumerate}
        \item We introduce a novel \textit{unified} framework within the transformer architecture that leverages \textbf{attention-} and \textbf{illuminance-maps} to enhance precision in processing affected regions at the pixel-level, addressing the challenges of underexposure, overexposure, and mixed exposure in images as a single task.

        \item Our method achieves remarkable efficiency, with an average inference speed of \textbf{0.095 seconds} per image\footnote{computed over LOL-v2 test dataset following previous benchmarks \cite{Cui_2022_BMVC,guo2020zero,chen2021pre,jiang2017learning}.}, significantly faster with lesser memory consumption than many existing frameworks. Coupled with a compact architecture of only \textbf{101 thousand} parameters, the model is ideal for deployment on edge devices.

        \item We present a novel ``\textbf{MUL-ADD}'' loss function that intelligently combines contrast scaling and brightness shifting to adaptively enhance images, improving dynamic range and preventing over-smoothing.

        \item We develop an Exposure-Aware Fusion (\textbf{EAF}) Block, designed for the efficient fusion of local and global features. This block refines image exposure corrections with heightened precision, enabling context-aware enhancements tailored to the specific exposure needs of each image region.
    \end{enumerate}

\section{Related Work}
\label{sec:related_work}
    
    \textbf{Traditional to Advanced Deep Learning Techniques.}\quad In the realm of image enhancement and exposure correction, significant strides have been made to address the challenges posed by exposure scenarios. Early techniques \cite{celik2011contextual,ibrahim2007brightness,lee2013contrast} leveraged contrast-based histogram equalization (HE), laying the groundwork for more advanced methods. These initial approaches were followed by studies in Retinex theory, which focused on decomposing images into reflection and illumination maps \cite{land1986alternative,land1977retinex}. The advent of deep learning transformed exposure correction, with a shift from enhancing low-light images to addressing both underexposure and overexposure \cite{afifi2020deep,afifi2021learning,chen2018deep,guo2020zero,Wang_2019_CVPR,xu2020learning,yu2018deepexposure,10.1145/3343031.3350926}. The notable work of Afifi et al. \cite{afifi2020deep} stands out, employing deep learning to simultaneously address underexposure and overexposure, a task not adequately considered by previous methodologies. There was a momentous shift towards convolutional neural network (CNN)-based methods, achieving state-of-the-art results and improving the accuracy and efficiency of exposure correction algorithms \cite{guo2020zero,kim2021representative,moran2020deeplpf,sharma2021nighttime,Wang_2019_CVPR,xu2022snr}.
    
    \vspace{.1in}\noindent\textbf{Addressing the Challenges of Mixed Exposure.}\quad Despite these advancements, the challenges of mixed exposure have remained relatively unaddressed in high-contrast scenarios. Benchmark datasets such as LOL \cite{WeiWY018}, LOL-[4K;8K] \cite{wang2023ultra}, SID \cite{ChenCXK18}, SICE \cite{Cai2018deep}, and ELD \cite{wei2020physics} offer limited mixed exposure instances, highlighting a gap in both data and models. Synthetic datasets like SICE-Grad and SICE-Mix are enabling the development of methods tailored to mixed exposure scenarios \cite{liu2021benchmarking,zheng2020optical,cui2021multitask,sasagawa2020yolo}. Some representative methods of deep learning like RetinexNet \cite{WeiWY018} and KIND \cite{10.1145/3343031.3350926}, focusing on illumination and reflectance component restoration in images, achieved good performance, but most methods emphasize either underexposure or overexposure correction, and fail to correct the combination. More recent studies have attempt to address the challenges of correcting both underexposed and overexposed images \cite{wang2022local,drbn}, a task complicated by the differing optimization processes required for each type of exposure. MSEC \cite{afifi2021learning}, a revolutionary work in this area, utilizes a Laplacian pyramid structure to incrementally restore brightness and details, dealing with a range of exposure levels. 

    To manage the correction of a wide range of exposures, several recent works were proposed, such as Huang et al.~\cite{huang2022exposure} using exposure normalization, CMEC \cite{nsampi2021learning} targeting exposure-invariant feature spaces for exposure consistency, and ECLNet \cite{huang2022exposure} using bilateral activation for exposure consistency. Moreover, Wang et al.~\cite{wang2022local} tackle the issue of uneven illumination, while FECNet \cite{huang2022deep} uses a Fourier-based approach for lightness and structure. Still challenges remain unsolved: (1) handling nonuniform illumination, (2) simultaneously addressing both overexposure and underexposure within the same frame, and, (3) ensuring that global adjustments do not adversely affect local regions. Our work addresses these challenges by introducing a unified framework that uses attention and illuminance maps to process mixed exposure regions more precisely. 
    
    \vspace{.1in}\noindent\textbf{Emerging Trends with Computational Challenges.}\quad Recent studies have begun addressing the dual challenge of correcting under and overexposed images through innovative architectures, including transformers \cite{dosovitskiy2020image}, despite their computational intensity as noted in works like Vaswani et al. \cite{NIPS2017_3f5ee243}. The realm of image enhancement has seen remarkable models with human-level enhancement capabilities, such as ExposureDiffusion \cite{wang2023exposurediffusion}, Diff-retinex \cite{yi2023diff}, wavelet-based diffusion \cite{jiang2023low}, PyDiff (Pyramid Diffusion) \cite{zhou2023pyramid}, Global structure aware diffusion \cite{hou2024global}, LLDiffusion \cite{wang2023lldiffusion}, and Maximal diffusion values \cite{kim2019low}, demonstrating significant advancements. However, these resource-intensive models, alongside the emerging vision-language models (VLMs) in image restoration, present deployment challenges on edge devices, often requiring tens of seconds to process a single HD or FHD image \cite{Wang_2023_CVPR,luo2023controlling}. In contrast, Unified-EGformer achieves an average inference speed of ${\sim}$200 milliseconds on HD images, $9.6\times$ faster on average inference time among representative models in Tab.~2 in Ye et al.~\cite{s24041345}.

\section{Methodology: Unified-EGformer}
\label{sec:method}
    \begin{figure*}[ht]
        \centering
        \includegraphics[width=0.95\textwidth]{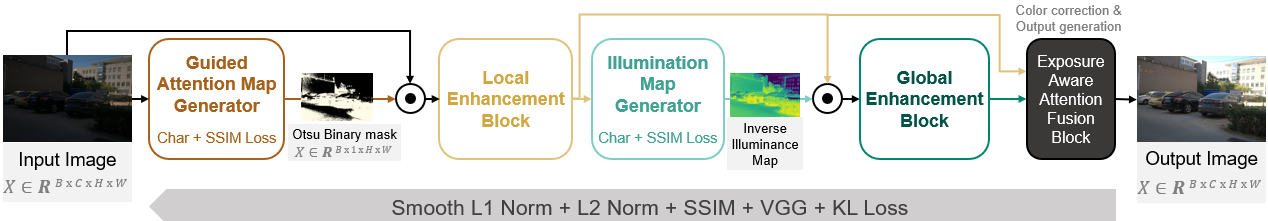} \\
        \includegraphics[width=0.94\textwidth]{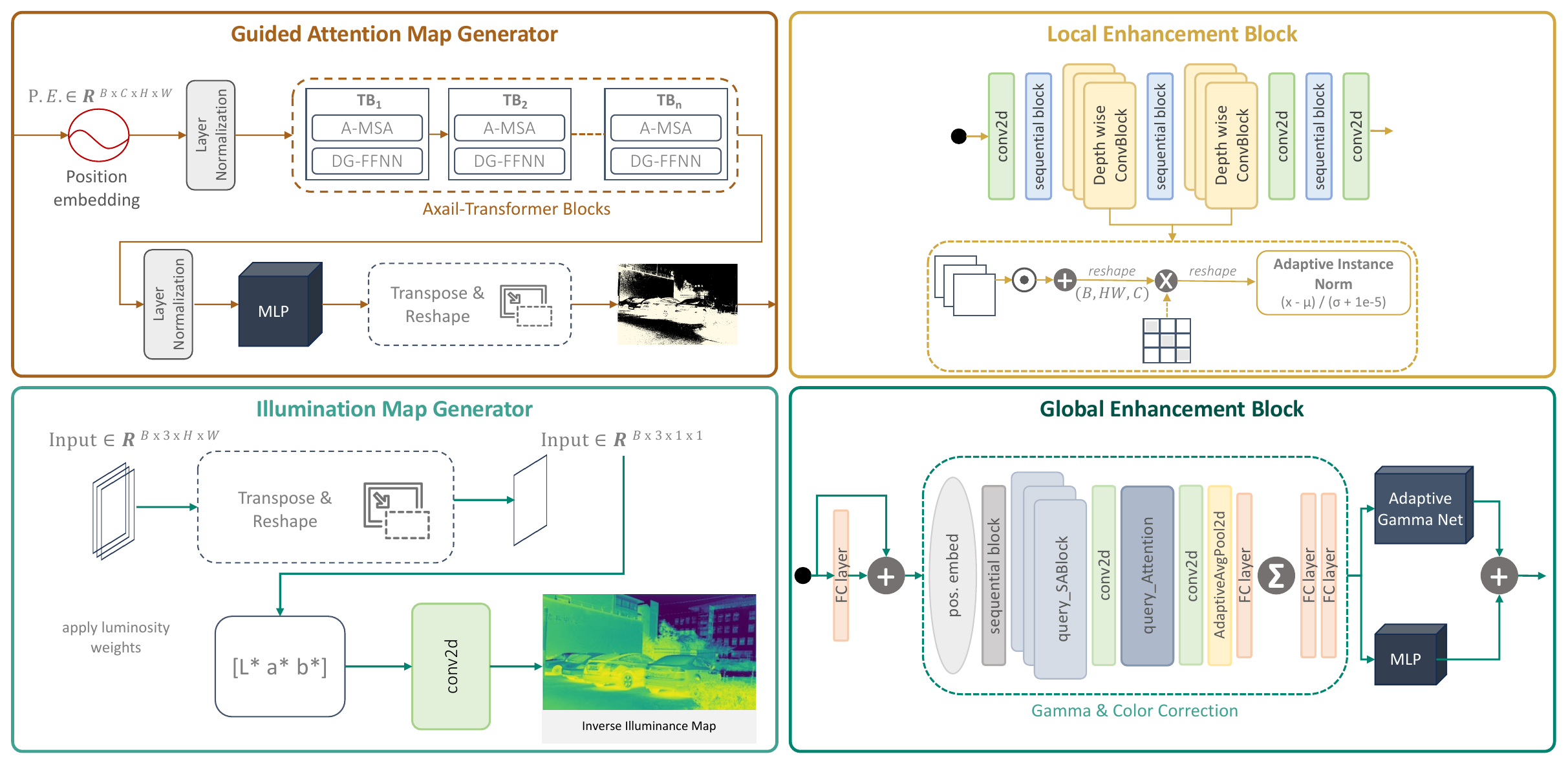}
        \caption{U-EGformer's training, fine-tuning and inference pipelines. All four modules are showcased: Guided Attention Map, Local Block, Global Block, and Exposure Aware Fusion block.}
        \label{fig:pipeline}
    \end{figure*}
    \vspace{-.05in}\noindent Unifed-EGformer achieves image enhancement through an Attention Map Generation mechanism that identifies exposure adjustment regions, a Local Enhancement Block for pixel-wise refinement, a Global Enhancement Block for color and contrast adjustment, and an Exposure Aware Fusion (EAF) block that fuses features from both enhancements for balanced exposure correction as shown in Fig.~\ref{fig:pipeline}.

    \subsection{Guided Map Generation}
    Unified-EGformer introduces significant advancements in the attention mechanism and feed-forward network within its architecture to adeptly handle mixed exposure scenarios. These enhancements are encapsulated as follows:
    
    \begin{figure}[ht]
        \centering
        \begin{subfigure}{.26\columnwidth}
            \centering
            \includegraphics[width=0.92\linewidth]{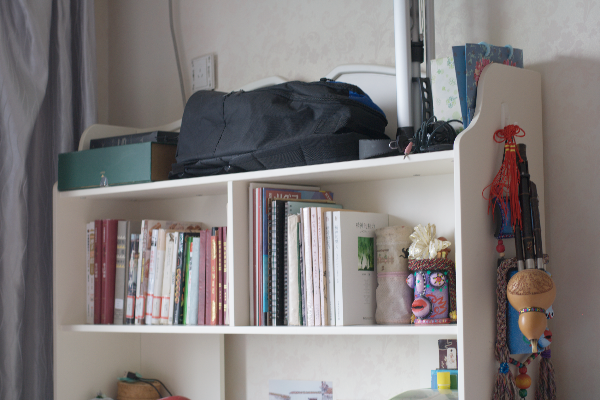}
            \caption{Input}
        \end{subfigure}
        \hfill 
        \begin{subfigure}{.26\columnwidth}
            \centering
            \includegraphics[width=0.92\linewidth]{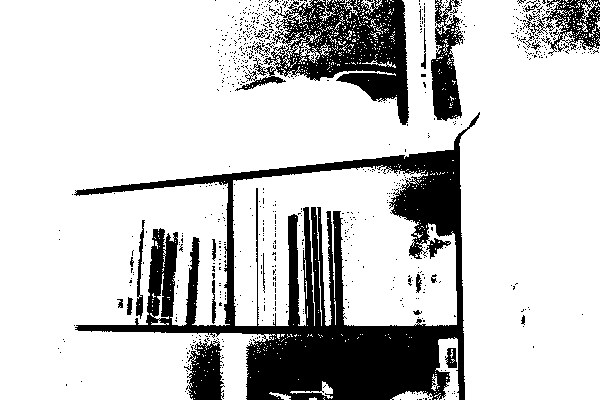}
            \caption{Single Exposure}
        \end{subfigure}%
        \hfill 
        \begin{subfigure}{.26\columnwidth}
            \centering
            \includegraphics[width=0.92\linewidth]{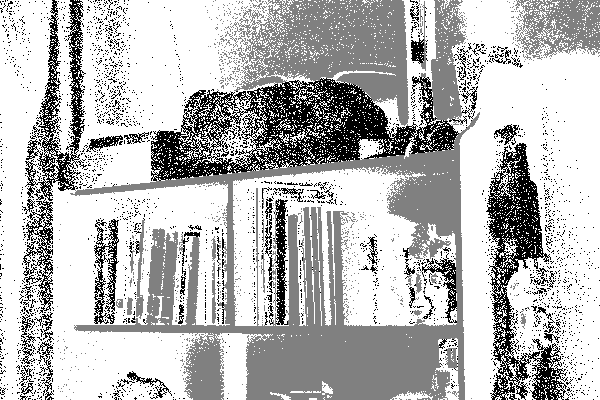}
            \caption{Bi-Exposure}
        \end{subfigure}
        \vspace{-2mm}
        \caption{Visualization of Otsu thresholding challenge: \textbf{(a)} original image, \textbf{(b)} mask for single exposure (underexposed), and \textbf{(c)} mask for bi-exposure (under and overexposed).\vspace{-.2in}
        }
        \label{fig:applying-otsu-directly}
    \end{figure}
    \begin{figure*}[tb]
        \centering
        \includegraphics[width=0.75\textwidth]{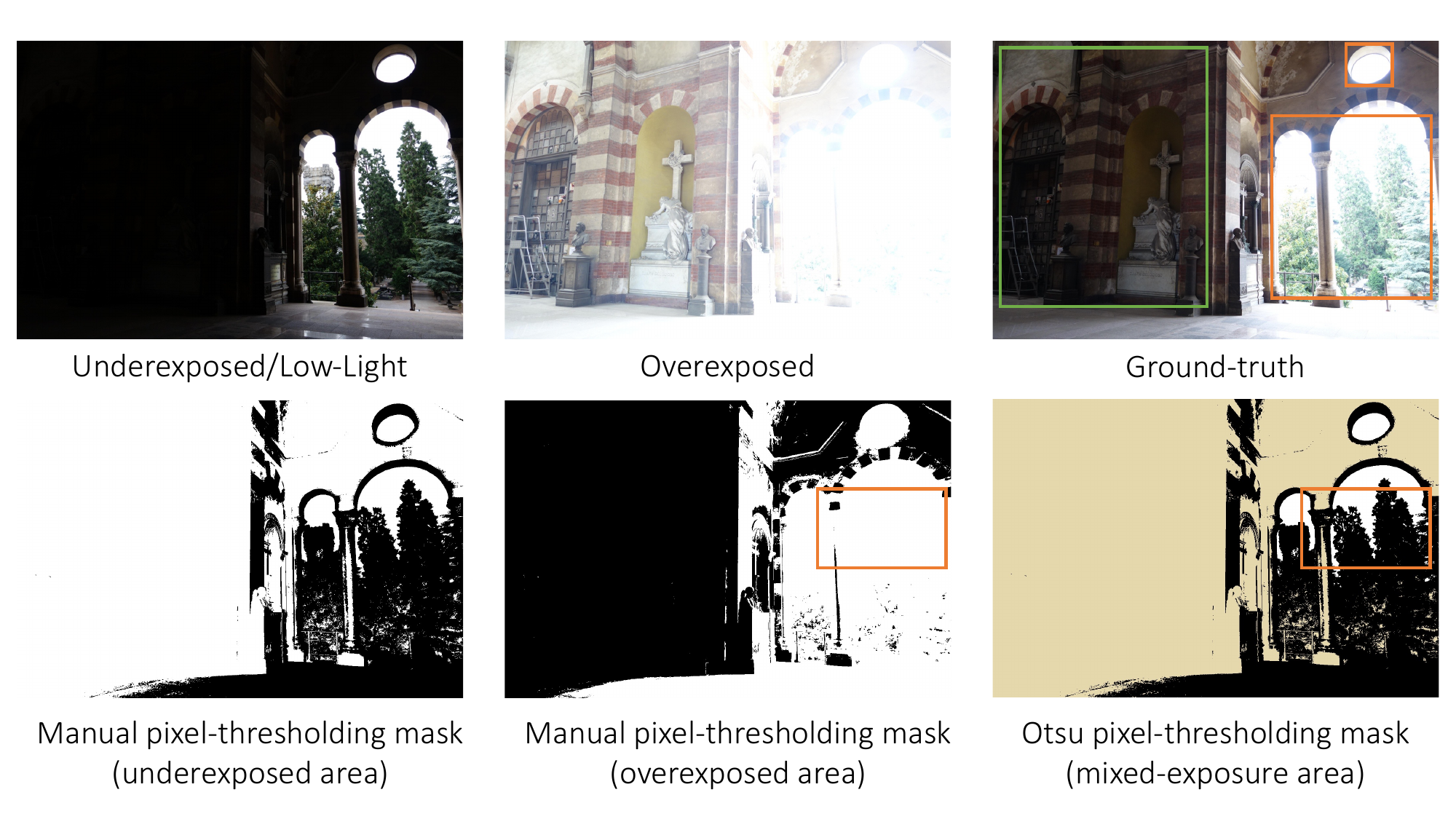} \vspace{-0.5em}
        \caption{The top row, from left to right: an underexposed image, an overexposed image, and the ground truth. The bottom row illustrates pixel-thresholding binary masks for the underexposed (white indicating underexposed regions), overexposed (white indicating overexposed regions) and Otsu thresholding for mixed exposures (yellow indicating underexposed regions, white representing overexposed areas, and black as correctly exposed portions.\vspace{-.2in}}
        \label{fig:attentionmap-thresholding}
    \end{figure*}
    \vspace{.1in}\noindent\textbf{Thresholding.}
    To highlight the sub-regions of the images with impacted exposure problems, we need a way to point out those impacted set of pixels within the input. We use Otsu thresholding \cite{otsu}, a traditional yet effective technique. It is a global thresholding technique for automatic thresholding that works by selecting the threshold to minimize \textit{intra-class variance} (variance within a class) or maximize \textit{inter-class variance} (variance between classes). However, this method induces granular noise in the image \cite{otsu} due to the non-uniform pixels in low lux regions. The noise is highlighted by the resultant masks as shown in Fig.~\ref{fig:applying-otsu-directly}, and will influence the subsequent exposure correction.

    To mitigate noise, we implemented adaptive thresholding using pixel blocks and downsampled images. We further reduced noise creep with nearest neighbor downsampling and Gaussian blur. Integrating Charbonnier loss \cite{charbonnier} into our attention map mechanism encouraged smoother transitions in areas of high gradient variance, specifically targeting denoising. This component, combined with the SSIM loss that is applied directly on the input, synergistically contributes to noise reduction.

    \begin{wrapfigure}[18]{L}{0.3\textwidth}
        \centering
        \vspace{-2em}
        \includegraphics[width=0.9\textwidth]{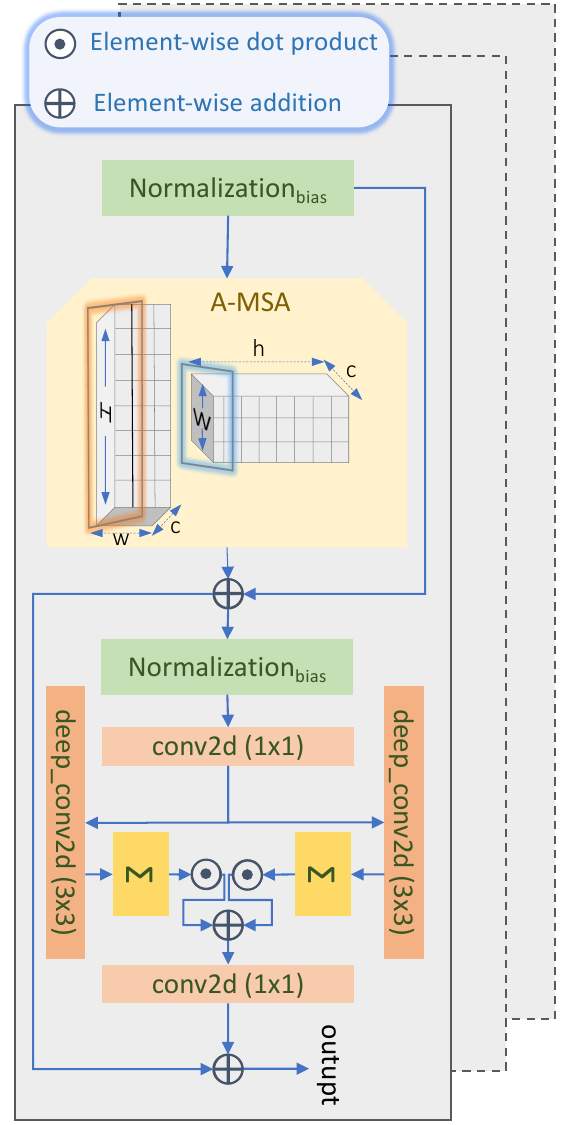}
        \caption{Chain of efficient transformer blocks equipped with A-MSA, DGFN.}
        \label{fig:trans-block}
    \end{wrapfigure}

    Our implementation of threshold selection is as follows. First, we calculate the Otsu average across the training set to establish a baseline for automatic thresholding. We then  apply this average threshold to each image in the dataset. Using a data set-specific threshold, this method ensures a more uniform application\footnote{reducing noise propensity in low-light conditions and enhancing exposure correction consistency} of the Otsu method.

    \vspace{.1in}\noindent\textbf{Attention Map Generator.}
    \label{sec:guided-attnetion-map-generator}
    The Unified-EGformer begins with a guided attention map generator, designed to identify regions within an image affected by mixed exposure. This process involves generating a map $\boldsymbol{M_g} \in \mathbb{R}^{B \times H \times W \times C}$, where $H$, $W$, and $C$ represent the height, width, and channel dimensions of the input image $x \in \mathbb{R}^{H \times W \times C}$. This map, $m \in \boldsymbol{M_g}$, is used in an element-wise dot product with the image, resulting in a guided input image that undergoes underexposed, overexposed, or mixed exposure enhancement, as depicted in Fig.~\ref{fig:attentionmap-thresholding}, demonstrating how we apply Otsu thresholding to get attention masks labels.

    The architecture incorporates improved Swin transformer \cite{Liu_2021_ICCV} blocks, leveraging Axis-based Multi-head Self-Attention (A-MSA) and Dual Gated Feedforward Network (DGFN) \cite{wang2023ultra,NIPS2017_3f5ee243,dosovitskiy2020image} for efficient, fast and focused feature processing. The A-MSA reduces computational load by applying self-attention across height and width axes sequentially, optimizing for local contexts within high-resolution images, as illustrated in Fig.~\ref{fig:trans-block} (see details in Supplement Section~7.2
    The DGFN introduces a dual gating mechanism to the feedforward network, allowing selective emphasis on critical features necessary to distinguish and correct underexposed and overexposed areas effectively (see Supplement Section~7.3).

    \vspace{.1in}\noindent\textbf{Illumination Map Generator.}
    \begin{figure*}[!tb]
        \centering
        \label{fig:illum_map}
        \includegraphics[width=0.9\textwidth]{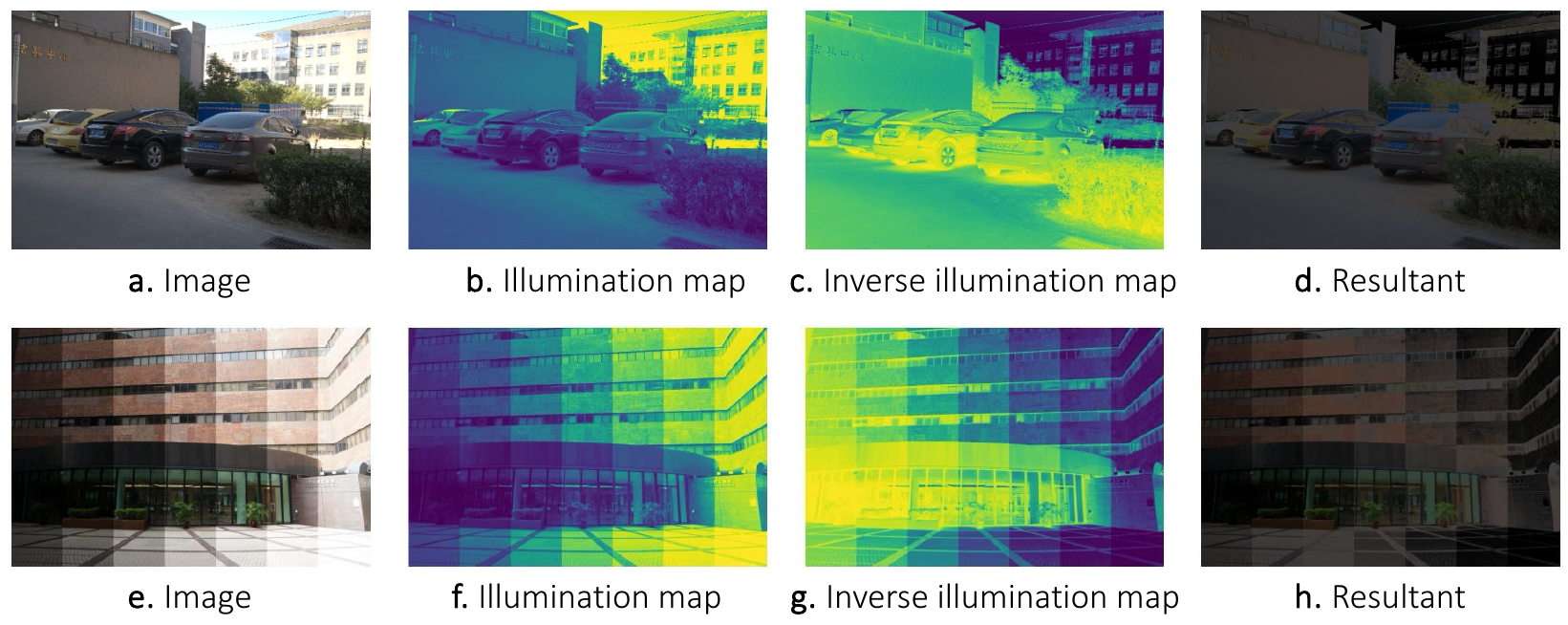} \vspace{-0.5em}
        \caption{Ground-truth \textbf{(a,e)} with it's corresponding illuminance map \textbf{[b,f]}, inverse illuminance maps \textbf{[c,g]}, and the enhanced results showcasing the effectiveness of the inverse illuminance normalization \textbf{[d,h]}. Note that the illuminance maps are using false color.\vspace{-.2in}}
    \end{figure*}
    We incorporate the generation of an illumination map $I_{\text{illum}}$ into the global enhancement block, providing a foundational layer for exposure correction. Unlike the complex attention mechanisms required for a local block, generating an illumination map leverages a direct conversion from RGB to luminance ($I_{\text{rgb}} \rightarrow I_{\text{illum}}$), offering a simpler and faster solution ($I_{\text{illum}} = \mathbf{W}\times I_{\text{rgb}}$), where $\mathbf{W}$ corresponds to the luminosity method \cite{series2011studio}. The block can utilize the illuminance information to dynamically adjust global parameters such as color balance and exposure levels, ensuring the enhancements are computationally efficient. Our ablation studies shown in Tab.~\ref{tab:ablation-study} confirm that these enhancements are perceptually meaningful compared to the baseline without the illumination map.

    \subsection{Unified-Enhancement}
    \noindent\textbf{Local Enhancement Block (LEB).}
    \label{sec:local_block}
    The LEB takes a dot product of attention map $\mathcal{A}(x)$ and the sRGB $I(x)$ image, that uniquely forms an objective mapping from input to output. The LEB applies a lightweight convolutional block inspired from PE-Yolo \cite{yin2023peyolopyramidenhancementnetwork}, UWFormer \cite{chen2024uwformerunderwaterimageenhancement} localized pixel-wise refinement. We utilize an adaptive instance-based color normalization for capturing a wider range of colors based on individual inputs. Unlike traditional methods, our approach maintains the integrity of features without down-sampling and up-sampling. The output of this block calculates multiplicative ($M_L$) and additive ($A_L$) correction factors through a feed-forward network, enabling precise pixel-wise enhancement. The local enhancement is formulated as:
    \begin{equation}
        \hat{I}(x) = M_L(x) \odot I(x) + A_L(x)
    \end{equation}
    where $\hat{I}(x)$ is the locally enhanced image, $I(x)$ the original image, and $\odot$ denotes element-wise multiplication.

    \vspace{8pt}
    \noindent \textbf{Global Enhancement Block (GEB).}
    \label{sec:global_block}
    Complementary to the LEB, the global enhancement block adjusts the image's overall exposure through adaptive gamma correction and color balance. Unlike static bias adjustments \cite{Cui_2022_BMVC}, this block dynamically calculates the gamma correction factor and color transformation parameters based on the image's content. 
    We implement a convolutional subnetwork with GELU activations and adaptive average pooling, followed by a sigmoid to automatically compute global parameters, denoted by $\theta$.
    This adaptive approach to global enhancement allows for a more nuanced and content-aware balance of contrast and color. The global correction function is described as:
    \begin{equation}
    \label{eq:global}
        \text{G}(I) = f_{\gamma}(I;\theta), \quad I \in \mathbb{R}^{H \times W \times C}
    \end{equation}
    where $G(I)$ is the globally enhanced image, and $f_{\gamma}$ represents the function for global adjustments, influenced by the calculated parameters $\theta$.

    \subsection{Exposure-Aware Fusion (EAF) Block} 
    \label{sec:eaf}
    Our novel exposure aware fusion block is architecturally designed to integrate local and global enhancement features, enabling comprehensive image enhancement. The fusion process begins with two convolutional layers that apply spatial filtering to extract the salient features necessary for exposure correction. We also use global average pooling, mapping the feature maps to the global context vector. These fusion weights serve as a gating mechanism to regulate the contribution of local and global features. They are adaptively learned, encapsulating both detailed texture information and broad illumination context.

    \subsection{Loss Functions}
    \label{sec:loss}
    To enhance multiple aspects of image quality, our training uses a detailed loss function setup in the RGB color space. It includes $L_1$ and $L_2$ losses for handling outliers and preserving fine details, SSIM for maintaining structural integrity, and VGG for ensuring semantic consistency. We also incorporate a novel MUL-ADD loss to accurately adjust the image's contrast and brightness, ensuring that the dynamic range is well represented without blurring details. The VGG loss helps match the output to high-level visual quality standards.

    \textbf{MUL-ADD loss.}\quad The Multiplicative-Additive loss is specifically designed to handle mixed exposure scenarios by optimizing the multiplicative and additive adjustments applied during the enhancement process. This loss is mathematically defined as:
    \begin{align}
        L_{\text{MA}}(M_L, A_L, I_{\text{low}}, I_{\text{high}}) = \boldsymbol{\xi}~L_1(M_L, \hat{M}(I_{\text{low}}, I_{\text{high}})) + \boldsymbol{\psi}~L_1(A_L, \hat{A}(I_{\text{low}}, I_{\text{high}}))
    \vspace{-2em}
    \end{align}
    Where:
    \begin{align}
    \vspace{-2em}
        \hat{M}(I_{\text{low}}, I_{\text{high}}) = \frac{I_{\text{high}} - \hat{A}(I_{\text{low}}, I_{\text{high}})}{I_{\text{low}} + \epsilon}, \quad \hat{A}(I_{\text{low}}, I_{\text{high}}) = I_{\text{high}} - \frac{I_{\text{high}} \cdot \epsilon}{I_{\text{low}} + \epsilon}
    \end{align}
    In these equations, $I_{\text{low}}$ and $I_{\text{high}}$ represent the low-light input and the target high-quality image, respectively, while $M_{\text{L}}$ and $A_{\text{L}}$ are the multiplicative and additive components learned by the local enhancement block. The parameters $\boldsymbol{\xi}$ and $\boldsymbol{\psi}$ control the balance between the two components of the Mul-Add loss. The small constant $\epsilon=1e-8$ is added to avoid division by zero. This loss function ensures that the multiplicative and additive factors are optimized to produce a natural-looking enhancement without overcompensating in either direction.
    Our combined loss function $\mathcal{\text{L}}_{\text{C}}$, considering both local and global outputs, is detailed below:
    \begin{equation}
    \label{eq:pretraining}
    \begin{split}
        \mathcal{L}_{\text{C}}(y, \hat{y}, I_{low}, I_{high}) = & \ \boldsymbol{\alpha}~L_{1}(y, \hat{y})_{(l,g)} + \boldsymbol{\beta}~L_2(y, \hat{y}) + \boldsymbol{\gamma}~L_{\text{SSIM}}(y, \hat{y}) + \boldsymbol{\delta}~L_{\text{VGG}}(y, \hat{y}) \\
        & + \boldsymbol{\eta}~L_{\text{MA}}(M_{L}, A_{L}, I_{\text{low}}, I_{\text{high}}) + L_{\text{attn}}(M_g(b), M)
    \end{split}
    \end{equation}
    where $\alpha, \beta, \gamma, \delta,$ and $\eta$ are hyperparameters balancing the influence of each loss term, $y$ is the ground truth, $\hat{y}$ is the predicted image, $M_{L}$ and $A_{L}$ are the multiplicative and additive components of the local block, $I_{low}$ is the low-light input, and $I_{high}$ is the target high-quality image. Our fine-tuning stage's loss equation can be presented with the physics-based KL-divergence loss:
    \begin{align}
    \label{eq:finetuning}
        \mathcal{L}_{\text{finetune}}(y, \hat{y}, P, Q) = & \ \boldsymbol{\lambda}~L_1(y, \hat{y}) + \boldsymbol{\mu}~L_{\text{SSIM}}(y, \hat{y}) + \boldsymbol{\nu}~L_{\text{KL}}(\mathcal{P}, Q)_{(l,g)}
    \end{align}
    %

\begin{table*}[!ht]
    \centering
    \caption{Results for our exposure guided transformer approach over ME-v2 \cite{afifi2021learning} and SICE-v2 \cite{Cai2018deep} datasets. \firsttone{}, \secondtone{}, \thirdtone{} denotes top three respectively. We did not include other recent models that are too complex ($>10M$ params).\vspace{-2em}}
    \label{tab:mev2_sicev2}
    \scalebox{0.63}{
    \begin{tabular}{|l|ccccccccccccc|}
    \hline
    \multicolumn{1}{|c}{\multirow{3}{*}{\textbf{Method}}} &
      \multicolumn{6}{|c}{\textbf{ME-v2}} &
      \multicolumn{6}{c|}{\textbf{SICE-v2}} &
      \multicolumn{1}{c|}{\multirow{3}{*}{\#params}} \\ \cline{2-13}
    \multicolumn{1}{|c}{} &
      \multicolumn{2}{|c}{Underexposure} &
      \multicolumn{2}{c}{Overexposure} &
      \multicolumn{2}{c|}{Average} &
      \multicolumn{2}{c}{Underexposure} &
      \multicolumn{2}{c}{Overexposure} &
      \multicolumn{2}{c|}{Average} &
      \multicolumn{1}{c|}{} \\ \cline{2-13}
    \multicolumn{1}{|c}{} &
      \multicolumn{1}{|c}{PSNR $\uparrow$} &
      \multicolumn{1}{c}{SSIM $\uparrow$} &
      \multicolumn{1}{c}{PSNR $\uparrow$} &
      \multicolumn{1}{c}{SSIM $\uparrow$} &
      \multicolumn{1}{c}{PSNR $\uparrow$} &
      \multicolumn{1}{c|}{SSIM $\uparrow$} &
      \multicolumn{1}{c}{PSNR $\uparrow$} &
      \multicolumn{1}{c}{SSIM $\uparrow$} &
      \multicolumn{1}{c}{PSNR $\uparrow$} &
      \multicolumn{1}{c}{SSIM $\uparrow$} &
      \multicolumn{1}{c}{PSNR $\uparrow$} &
      \multicolumn{1}{c|}{SSIM $\uparrow$} &
      \multicolumn{1}{c|}{} \\ \hline

    RetinexNet \cite{WeiWY018} &
        \multicolumn{1}{c}{12.13} &
        \multicolumn{1}{c}{0.6209} &
        \multicolumn{1}{c}{10.47} &
        \multicolumn{1}{c}{0.5953} &
        \multicolumn{1}{c}{11.14} &
        \multicolumn{1}{c|}{0.6048} &
        \multicolumn{1}{c}{12.94} &
        \multicolumn{1}{c}{0.5171} &
        \multicolumn{1}{c}{12.87} &
        \multicolumn{1}{c}{0.5252} &
        \multicolumn{1}{c}{12.90} &
        \multicolumn{1}{c|}{0.5212} &
        \multicolumn{1}{c|}{0.840M} \\

    URetinexNet \cite{9879970} &
        \multicolumn{1}{c}{13.85} &
        \multicolumn{1}{c}{0.7371} &
        \multicolumn{1}{c}{9.81} &
        \multicolumn{1}{c}{0.6733} &
        \multicolumn{1}{c}{11.42} &
        \multicolumn{1}{c|}{0.6988} &
        \multicolumn{1}{c}{12.39} &
        \multicolumn{1}{c}{0.5444} &
        \multicolumn{1}{c}{7.40} &
        \multicolumn{1}{c}{0.4543} &
        \multicolumn{1}{c}{12.40} &
        \multicolumn{1}{c|}{0.5496} &
        \multicolumn{1}{c|}{1.320M} \\

    Zero-DCE \cite{guo2020zero} &
        \multicolumn{1}{c}{14.55} &
        \multicolumn{1}{c}{0.5887} &
        \multicolumn{1}{c}{10.40} &
        \multicolumn{1}{c}{0.5142} &
        \multicolumn{1}{c}{12.06} &
        \multicolumn{1}{c|}{0.5441} &
        \multicolumn{1}{c}{16.92} &
        \multicolumn{1}{c}{0.6330} &
        \multicolumn{1}{c}{7.11} &
        \multicolumn{1}{c}{0.4292} &
        \multicolumn{1}{c}{12.02} &
        \multicolumn{1}{c|}{0.5211} &
        \multicolumn{1}{c|}{0.079M} \\

    Zero-DCE++ \cite{Zero-DCE++} &
        \multicolumn{1}{c}{13.82} &
        \multicolumn{1}{c}{0.5887} &
        \multicolumn{1}{c}{9.74} &
        \multicolumn{1}{c}{0.5142} &
        \multicolumn{1}{c}{11.37} &
        \multicolumn{1}{c|}{0.5583} &
        \multicolumn{1}{c}{11.93} &
        \multicolumn{1}{c}{0.4755} &
        \multicolumn{1}{c}{6.88} &
        \multicolumn{1}{c}{0.4088} &
        \multicolumn{1}{c}{9.41} &
        \multicolumn{1}{c|}{0.4422} &
        \multicolumn{1}{c|}{0.010M} \\

    DPED \cite{ignatov2017dslr} &
        \multicolumn{1}{c}{20.06} &
        \multicolumn{1}{c}{0.6826} &
        \multicolumn{1}{c}{13.14} &
        \multicolumn{1}{c}{0.5812} &
        \multicolumn{1}{c}{15.91} &
        \multicolumn{1}{c|}{0.6219} &
        \multicolumn{1}{c}{16.83} &
        \multicolumn{1}{c}{0.6133} &
        \multicolumn{1}{c}{7.99} &
        \multicolumn{1}{c}{0.4300} &
        \multicolumn{1}{c}{12.41} &
        \multicolumn{1}{c|}{0.5217} &
        \multicolumn{1}{c|}{0.390M} \\

    KIND \cite{10.1145/3343031.3350926} &
        \multicolumn{1}{c}{15.51} &
        \multicolumn{1}{c}{0.7115} &
        \multicolumn{1}{c}{11.66} &
        \multicolumn{1}{c}{0.7300} &
        \multicolumn{1}{c}{13.20} &
        \multicolumn{1}{c|}{0.7200} &
        \multicolumn{1}{c}{15.03} &
        \multicolumn{1}{c}{0.6700} &
        \multicolumn{1}{c}{12.67} &
        \multicolumn{1}{c}{0.6700} &
        \multicolumn{1}{c}{13.85} &
        \multicolumn{1}{c|}{0.6700} &
        \multicolumn{1}{c|}{0.590M} \\

    DeepUPE \cite{Wang_2019_CVPR} &
        \multicolumn{1}{c}{19.10} &
        \multicolumn{1}{c}{0.7321} &
        \multicolumn{1}{c}{14.69} &
        \multicolumn{1}{c}{0.7011} &
        \multicolumn{1}{c}{16.25} &
        \multicolumn{1}{c|}{0.7158} &
        \multicolumn{1}{c}{16.21} &
        \multicolumn{1}{c}{0.6807} &
        \multicolumn{1}{c}{11.98} &
        \multicolumn{1}{c}{0.5967} &
        \multicolumn{1}{c}{14.10} &
        \multicolumn{1}{c|}{0.6387} &
        \multicolumn{1}{c|}{7.790M} \\

    SID \cite{ChenCXK18} &
        \multicolumn{1}{c}{19.37} &
        \multicolumn{1}{c}{0.8103} &
        \multicolumn{1}{c}{18.83} &
        \multicolumn{1}{c}{0.8055} &
        \multicolumn{1}{c}{19.04} &
        \multicolumn{1}{c|}{0.8074} &
        \multicolumn{1}{c}{19.51} &
        \multicolumn{1}{c}{0.6635} &
        \multicolumn{1}{c}{16.79} &
        \multicolumn{1}{c}{0.6444} &
        \multicolumn{1}{c}{18.15} &
        \multicolumn{1}{c|}{0.6540} &
        \multicolumn{1}{c|}{7.760M} \\

    SID-ENC \cite{huang2022exposure} &
        \multicolumn{1}{c}{22.59} &
        \multicolumn{1}{c}{0.8423} &
        \multicolumn{1}{c}{22.36} &
        \multicolumn{1}{c}{0.8519} &
        \multicolumn{1}{c}{22.45} &
        \multicolumn{1}{c|}{0.8481} &
        \multicolumn{1}{c}{21.36} &
        \multicolumn{1}{c}{0.6652} &
        \multicolumn{1}{c}{19.38} &
        \multicolumn{1}{c}{0.6843} &
        \multicolumn{1}{c}{20.37} &
        \multicolumn{1}{c|}{0.6748} &
        \multicolumn{1}{c|}{>7.760M} \\

    RUAS \cite{ll_benchmark} &
        \multicolumn{1}{c}{13.43} &
        \multicolumn{1}{c}{0.6807} &
        \multicolumn{1}{c}{6.39} &
        \multicolumn{1}{c}{0.4655} &
        \multicolumn{1}{c}{9.20} &
        \multicolumn{1}{c|}{0.5515} &
        \multicolumn{1}{c}{16.63} &
        \multicolumn{1}{c}{0.5589} &
        \multicolumn{1}{c}{4.54} &
        \multicolumn{1}{c}{0.3196} &
        \multicolumn{1}{c}{10.59} &
        \multicolumn{1}{c|}{0.4394} &
        \multicolumn{1}{c|}{0.002M} \\

    SCI \cite{Ma_2022_CVPR} &
        \multicolumn{1}{c}{9.96} &
        \multicolumn{1}{c}{0.6681} &
        \multicolumn{1}{c}{5.83} &
        \multicolumn{1}{c}{0.5190} &
        \multicolumn{1}{c}{7.49} &
        \multicolumn{1}{c|}{0.5786} &
        \multicolumn{1}{c}{17.86} &
        \multicolumn{1}{c}{0.6401} &
        \multicolumn{1}{c}{4.45} &
        \multicolumn{1}{c}{0.3629} &
        \multicolumn{1}{c}{12.49} &
        \multicolumn{1}{c|}{0.5051} &
        \multicolumn{1}{c|}{0.001M} \\

    MSEC \cite{afifi2021learning} &
        \multicolumn{1}{c}{20.52} &
        \multicolumn{1}{c}{0.8129} &
        \multicolumn{1}{c}{19.79} &
        \multicolumn{1}{c}{0.8156} &
        \multicolumn{1}{c}{20.08} &
        \multicolumn{1}{c|}{0.8210} &
        \multicolumn{1}{c}{19.62} &
        \multicolumn{1}{c}{0.6512} &
        \multicolumn{1}{c}{17.59} &
        \multicolumn{1}{c}{0.6560} &
        \multicolumn{1}{c}{18.58} &
        \multicolumn{1}{c|}{0.6536} &
        \multicolumn{1}{c|}{7.040M} \\

    CMEC \cite{nsampi2021learning} &
        \multicolumn{1}{c}{22.23} &
        \multicolumn{1}{c}{0.8140} &
        \multicolumn{1}{c}{22.75} &
        \multicolumn{1}{c}{0.8336} &
        \multicolumn{1}{c}{22.54} &
        \multicolumn{1}{c|}{0.8257} &
        \multicolumn{1}{c}{17.68} &
        \multicolumn{1}{c}{0.6592} &
        \multicolumn{1}{c}{18.17} &
        \multicolumn{1}{c}{0.6811} &
        \multicolumn{1}{c}{17.93} &
        \multicolumn{1}{c|}{0.6702} &
        \multicolumn{1}{c|}{5.400M} \\

    LCDPNet \cite{wang2022local} &
        \multicolumn{1}{c}{22.35} &
        \multicolumn{1}{c}{0.8650} &
        \multicolumn{1}{c}{22.17} &
        \multicolumn{1}{c}{0.8476} &
        \multicolumn{1}{c}{22.30} &
        \multicolumn{1}{c|}{0.8552} &
        \multicolumn{1}{c}{17.45} &
        \multicolumn{1}{c}{0.5622} &
        \multicolumn{1}{c}{17.04} &
        \multicolumn{1}{c}{0.6463} &
        \multicolumn{1}{c}{17.25} &
        \multicolumn{1}{c|}{0.6043} &
        \multicolumn{1}{c|}{0.960M} \\

    DRBN \cite{drbn} &
        \multicolumn{1}{c}{19.74} &
        \multicolumn{1}{c}{0.8290} &
        \multicolumn{1}{c}{19.37} &
        \multicolumn{1}{c}{0.8321} &
        \multicolumn{1}{c}{19.52} &
        \multicolumn{1}{c|}{0.8309} &
        \multicolumn{1}{c}{17.96} &
        \multicolumn{1}{c}{0.6767} &
        \multicolumn{1}{c}{17.33} &
        \multicolumn{1}{c}{0.6828} &
        \multicolumn{1}{c}{17.65} &
        \multicolumn{1}{c|}{0.6798} &
        \multicolumn{1}{c|}{0.530M} \\

    DRBN+ERL \cite{huang2023learning} &
        \multicolumn{1}{c}{19.91} &
        \multicolumn{1}{c}{0.8305} &
        \multicolumn{1}{c}{19.60} &
        \multicolumn{1}{c}{0.8384} &
        \multicolumn{1}{c}{19.73} &
        \multicolumn{1}{c|}{0.8355} &
        \multicolumn{1}{c}{18.09} &
        \multicolumn{1}{c}{0.6735} &
        \multicolumn{1}{c}{17.93} &
        \multicolumn{1}{c}{0.6866} &
        \multicolumn{1}{c}{18.01} &
        \multicolumn{1}{c|}{0.6796} &
        \multicolumn{1}{c|}{0.530M} \\

    DRBN-ERL+ENC \cite{huang2023learning} &
        \multicolumn{1}{c}{22.61} &
        \multicolumn{1}{c}{0.8578} &
        \multicolumn{1}{c}{22.45} &
        \multicolumn{1}{c}{\thirdtone{0.8724}} &
        \multicolumn{1}{c}{22.53} &
        \multicolumn{1}{c|}{\secondtone{0.8651}} &
        \multicolumn{1}{c}{22.06} &
        \multicolumn{1}{c}{\thirdtone{0.7053}} &
        \multicolumn{1}{c}{19.50} &
        \multicolumn{1}{c}{\firsttone{0.7205}} &
        \multicolumn{1}{c}{20.78} &
        \multicolumn{1}{c|}{\secondtone{0.7129}} &
        \multicolumn{1}{c|}{0.580M} \\

    ELCNet \cite{huang2017arbitrary} &
        \multicolumn{1}{c}{22.37} &
        \multicolumn{1}{c}{0.8566} &
        \multicolumn{1}{c}{22.70} &
        \multicolumn{1}{c}{0.8673} &
        \multicolumn{1}{c}{22.57} &
        \multicolumn{1}{c|}{\thirdtone{0.8619}} &
        \multicolumn{1}{c}{22.05} &
        \multicolumn{1}{c}{0.6893} &
        \multicolumn{1}{c}{19.25} &
        \multicolumn{1}{c}{0.6872} &
        \multicolumn{1}{c}{20.65} &
        \multicolumn{1}{c|}{0.6861} &
        \multicolumn{1}{c|}{0.018M} \\

    IAT \cite{Cui_2022_BMVC} &
        \multicolumn{1}{c}{20.34} &
        \multicolumn{1}{c}{0.8440} &
        \multicolumn{1}{c}{21.47} &
        \multicolumn{1}{c}{0.8518} &
        \multicolumn{1}{c}{20.91} &
        \multicolumn{1}{c|}{0.8479} &
        \multicolumn{1}{c}{21.41} &
        \multicolumn{1}{c}{0.6601} &
        \multicolumn{1}{c}{\firsttone{22.29}} &
        \multicolumn{1}{c}{0.6813} &
        \multicolumn{1}{c}{\secondtone{21.85}} &
        \multicolumn{1}{c|}{0.6707} &
        \multicolumn{1}{c|}{0.090M} \\

    ELCNet+ERL \cite{huang2023learning} &
        \multicolumn{1}{c}{22.48} &
        \multicolumn{1}{c}{0.8424} &
        \multicolumn{1}{c}{22.58} &
        \multicolumn{1}{c}{0.8667} &
        \multicolumn{1}{c}{22.53} &
        \multicolumn{1}{c|}{0.8545} &
        \multicolumn{1}{c}{\thirdtone{22.14}} &
        \multicolumn{1}{c}{0.6908} &
        \multicolumn{1}{c}{19.47} &
        \multicolumn{1}{c}{0.6982} &
        \multicolumn{1}{c}{20.81} &
        \multicolumn{1}{c|}{0.6945} &
        \multicolumn{1}{c|}{0.018M} \\

    FECNet \cite{10.1145/3343031.3350855} &
        \multicolumn{1}{c}{22.19} &
        \multicolumn{1}{c}{0.8562} &
        \multicolumn{1}{c}{\firsttone{23.22}} &
        \multicolumn{1}{c}{\secondtone{0.8748}} &
        \multicolumn{1}{c}{22.70} &
        \multicolumn{1}{c|}{0.8655} &
        \multicolumn{1}{c}{22.01} &
        \multicolumn{1}{c}{0.6737} &
        \multicolumn{1}{c}{19.91} &
        \multicolumn{1}{c}{0.6961} &
        \multicolumn{1}{c}{20.96} &
        \multicolumn{1}{c|}{0.6849} &
        \multicolumn{1}{c|}{0.150M} \\

    FECNet+ERL \cite{huang2023learning} &
        \multicolumn{1}{c}{\firsttone{23.10}} &
        \multicolumn{1}{c}{\firsttone{0.8639}} &
        \multicolumn{1}{c}{\secondtone{23.18}} &
        \multicolumn{1}{c}{\firsttone{0.8759}} &
        \multicolumn{1}{c}{\firsttone{23.15}} &
        \multicolumn{1}{c|}{\firsttone{0.8711}} &
        \multicolumn{1}{c}{\secondtone{22.35}} &
        \multicolumn{1}{c}{0.6671} &
        \multicolumn{1}{c}{\thirdtone{20.10}} &
        \multicolumn{1}{c}{0.6891} &
        \multicolumn{1}{c}{\thirdtone{21.22}} &
        \multicolumn{1}{c|}{0.6781} &
        \multicolumn{1}{c|}{>0.150M} \\ \hline

    \rowcolor{gray!20}$\text{U-EGformer}$ &
        \multicolumn{1}{c}{\thirdtone{22.50}} &
        \multicolumn{1}{c}{0.8469} &
        \multicolumn{1}{c}{22.70} &
        \multicolumn{1}{c}{0.8510} &
        \multicolumn{1}{c}{\thirdtone{22.60}} &
        \multicolumn{1}{c|}{0.8490} &
        \multicolumn{1}{c}{21.63} &
        \multicolumn{1}{c}{\secondtone{0.7112}} &
        \multicolumn{1}{c}{19.74} &
        \multicolumn{1}{c}{\thirdtone{0.7046}} &
        \multicolumn{1}{c}{20.69} &
        \multicolumn{1}{c|}{\thirdtone{0.7079}} &
        \multicolumn{1}{c|}{0.099M} \\

    \rowcolor{gray!20}$\text{U-EGformer}_{\text{eaf}}$ &
        \multicolumn{1}{c}{\secondtone{22.82}} &
        \multicolumn{1}{c}{\secondtone{0.8578}} &
        \multicolumn{1}{c}{\thirdtone{22.90}} &
        \multicolumn{1}{c}{0.8558} &
        \multicolumn{1}{c}{\secondtone{22.86}} &
        \multicolumn{1}{c|}{0.8568} &
        \multicolumn{1}{c}{\firsttone{22.98}} &
        \multicolumn{1}{c}{\firsttone{0.7192}} &
        \multicolumn{1}{c}{\secondtone{21.84}} &
        \multicolumn{1}{c}{\secondtone{0.7102}} &
        \multicolumn{1}{c}{\firsttone{22.41}} &
        \multicolumn{1}{c|}{\firsttone{0.7179}} &
        \multicolumn{1}{c|}{0.102M} \\ \hline
    \end{tabular}
}
\end{table*}
\begin{table}[!htb]
    \caption{Experimental results for quantitative comparison of our proposed exposure guided transformer across various datasets.}
    \begin{subtable}{.5\linewidth}
        \centering
        \caption{Results on LOL-v1 \cite{WeiWY018}, LOL-v2 \cite{drbn}, Adobe FiveK \cite{fivek}}
        \label{tab:lolv1_lolv2_FiveK}
        \scalebox{0.6}{
        \begin{tabular}{|l|cc|cc|cc|}
        \hline
        \multicolumn{1}{|c|}{\multirow{2}{*}{\textbf{Methods}}} &
          \multicolumn{2}{c}{\textbf{LOL-v1}} &
          \multicolumn{2}{c}{\textbf{LOL-v2}} &
          \multicolumn{2}{c|}{\textbf{MIT-FiveK}} \\ \cline{2-7}
        \multicolumn{1}{|c|}{} &
          \multicolumn{1}{|c}{PSNR $\uparrow$} & SSIM $\uparrow$ &
          \multicolumn{1}{c}{PSNR $\uparrow$} & SSIM $\uparrow$ &
          \multicolumn{1}{c}{PSNR $\uparrow$} & SSIM $\uparrow$ \\
          \hline
        DRBN \cite{drbn} & \multicolumn{1}{c}{19.55} & 0.746 & \multicolumn{1}{c}{20.13} & \thirdtone{0.820} & \multicolumn{1}{c}{-} & - \\

        DPE & \multicolumn{1}{c}{-} & - & \multicolumn{1}{c}{-} & - & \multicolumn{1}{c}{23.80} & 0.880 \\

        Deep-UPE \cite{Wang_2019_CVPR} & \multicolumn{1}{c}{-} & - & \multicolumn{1}{c}{13.27} & 0.452 & \multicolumn{1}{c}{23.04} & 0.893 \\

        3D-LUT \cite{zeng2020lut} & \multicolumn{1}{c}{16.35} & 0.585 & \multicolumn{1}{c}{17.59} & 0.721 & \multicolumn{1}{c}{\secondtone{25.21}} & \secondtone{0.922} \\

        DRBN+ERL \cite{huang2023learning} & \multicolumn{1}{c}{19.84} & 0.824 & \multicolumn{1}{c}{-} & - & \multicolumn{1}{c}{22.14} & 0.873 \\

        ECLNet+ERL \cite{huang2023learning} & \multicolumn{1}{c}{22.01} & 0.827 & \multicolumn{1}{c}{-} & - & \multicolumn{1}{c}{23.71} & 0.853 \\

        FECNet+ERL \cite{huang2023learning} & \multicolumn{1}{c}{21.08} & \thirdtone{0.829} & \multicolumn{1}{c}{-} & - & \multicolumn{1}{c}{24.18} & 0.864 \\

        RetinexNet \cite{WeiWY018} & \multicolumn{1}{c}{16.77} & 0.462 & \multicolumn{1}{c}{18.37} & 0.723 & \multicolumn{1}{c}{-} & - \\

        KinD++ \cite{kind_plus} & \multicolumn{1}{c}{21.30} & 0.823 & \multicolumn{1}{c}{19.08} & 0.817 & \multicolumn{1}{c}{-} & - \\

        ElightenGAN \cite{jiang2021enlightengan} & \multicolumn{1}{c}{17.483} & 0.652 & \multicolumn{1}{c}{18.64} & 0.677 & \multicolumn{1}{c}{-} & - \\

        IAT \cite{Cui_2022_BMVC} & \multicolumn{1}{c}{23.38} & 0.809 & \multicolumn{1}{c}{\firsttone{23.50}} & \thirdtone{0.824} & \multicolumn{1}{c}{\secondtone{25.32}} & \secondtone{0.920} \\

        LLFormer \cite{wang2023ultra} & \multicolumn{1}{c}{\firsttone{25.75}} & 0.823 & \multicolumn{1}{c}{\firsttone{26.19}} & 0.819 & \multicolumn{1}{c}{-} & - \\

        MIRNet \cite{Zamir2020MIRNet} & \multicolumn{1}{c}{\secondtone{24.10}} & \secondtone{0.832} & \multicolumn{1}{c}{20.35} & 0.782 & \multicolumn{1}{c}{-} & - \\ \hline

        \rowcolor{gray!20}U-EGformer & \multicolumn{1}{c}{\thirdtone{23.56}} & \firsttone{0.836} & \multicolumn{1}{c}{\thirdtone{22.05}} & \firsttone{0.841} & \multicolumn{1}{c}{\thirdtone{24.89}} & \firsttone{0.928} \\ \hline
        \end{tabular}
        }
    \end{subtable}
    \begin{subtable}{.49\linewidth}
        \centering
        \caption{Results on SICE Grad and SICE Mix datasets \cite{zheng2022low}.}
        \label{tab:sicegrad_sice_mix}
        \scalebox{0.6}{
        \begin{tabular}{|l|ccc|ccc|}
        \hline
        \multicolumn{1}{|c}{\multirow{2}{*}{\textbf{Methods}}} &
          \multicolumn{3}{|c}{\textbf{SICE Grad}} &
          \multicolumn{3}{c|}{\textbf{SICE Mix}} \\ 
        \cline{2-7} 
        \multicolumn{1}{|c}{} &
          \multicolumn{1}{|c}{PSNR $\uparrow$} &
          \multicolumn{1}{c}{SSIM $\uparrow$} &
          \multicolumn{1}{c|}{LPIPS $\downarrow$} &
          \multicolumn{1}{c}{PSNR $\uparrow$} &
          \multicolumn{1}{c}{SSIM $\uparrow$} &
          \multicolumn{1}{c|}{LPIPS $\downarrow$} \\ 
        \hline
        RetinexNet \cite{WeiWY018} & 12.397 & 0.606 & 0.407 & 12.450 & 0.619 & 0.364 \\
        ZeroDCE \cite{guo2020zero} & 12.428 & 0.633 & 0.362 & 12.475 & 0.644 & 0.314 \\
        RAUS \cite{liu2021ruas}    & 0.864  & 0.493 & 0.525 & 0.8628 & 0.494 & 0.499 \\
        SGZ \cite{zheng2022semantic} & 10.866 & 0.607 & 0.415 & 10.987 & 0.621 & 0.364 \\
        LLFlow \cite{wang2021low} & 12.737 & 0.617 & 0.388 & 12.737 & 0.617 & 0.388 \\
        URetinexNet \cite{9879970} & 10.903 & 0.600 & 0.402 & 10.894 & 0.610 & 0.356 \\
        SCI \cite{ma2022toward} & 8.644  & 0.529 & 0.511 & 8.559  & 0.532 & 0.484 \\
        KinD \cite{10.1145/3343031.3350926} & 12.986 & \thirdtone{0.656} & 0.346 & 13.144 & \secondtone{0.668} & 0.302 \\
        KinD++ \cite{kind_plus} & \thirdtone{13.196} & \secondtone{0.657} & \thirdtone{0.334} & \thirdtone{13.235} & \thirdtone{0.666} & \thirdtone{0.295} \\ \hline

        \rowcolor{gray!20}U-EGformer
            & \secondtone{13.272}
            & 0.643 
            & \secondtone{0.273}
            & \secondtone{14.235}
            & 0.652
            & \secondtone{0.281} \\

        \rowcolor{gray!20}U-EGformer (\textit{finetuned}) 
            & \firsttone{14.724} 
            & \firsttone{0.665}
            & \firsttone{0.269}
            & \firsttone{15.101} 
            & \firsttone{0.670}
            & \firsttone{0.260} \\
        \hline
        \end{tabular}
        }
    \end{subtable}
\end{table}

\section{Experiments}
\label{experiments}

    \subsection{Framework Setting}

    \textbf{Datasets.} We employ eight diverse datasets to  train and evaluate our proposed model: LOL-v1~\cite{WeiWY018} and LOL-v2 for foundational training and testing with real-world and synthetic images; Multiple-Exposure ME-v2 tailored for diverse exposure scenarios; SICE, including the SICE-Grad and SICE-Mix subsets for gradient and mixed-exposure challenges, respectively; and MIT-FiveK for benchmarking against professionally retouched images. LOL-v1 contains 500 image pairs with 485 and 15 for training and testing datasets, where each image has a resolution of ($3\times600\times400$). LOL-v2 is divided into real and synthetic subsets with detailed configurations for training/testing (with 689 and 100 images for real-world); in the BAcklit Image Dataset (BAID) dataset \cite{lv2022backlitnet}, we only use 380 randomly selected training images from Liang et al.~\cite{liang2023iterative} and utilize the complete 368 2K resolution images from the test set.

    \vspace{8pt}
    \noindent\textbf{Training Strategy.} ~In tackling the mixed exposure challenge in image processing, our approach adopts a \textit{pre-training} stage and a \textit{finetuning} stage as shown in Fig.~\ref{fig:pipeline}. We engage in \textit{pre-training} using our custom loss function, $\mathcal{L}_{C}$ (Eq.~\ref{eq:pretraining}), which combines several loss components with individually set hyperparameters for input-output pairs. In the \textit{finetuning} phase, we refine the model with a physics-based pixel-wise reconstruction loss function tailored to camera sensors obeying Poisson distribution $\mathcal{P}$ \cite{wang2023exposurediffusion} (more details are in the supplementary material in Section~7.4).

    Both stages of training leverage a combination of loss functions, which are detailed in the following subsection. This systematic progression from foundational learning to focused refinement helps to address the complexities inherent in mixed exposure challenges. (More details can be found in the supplementary material.)

    \subsection{Qualitative Results} \vspace{-0.5em}

    \begin{figure*}
        \centering
        \includegraphics[width=\textwidth]{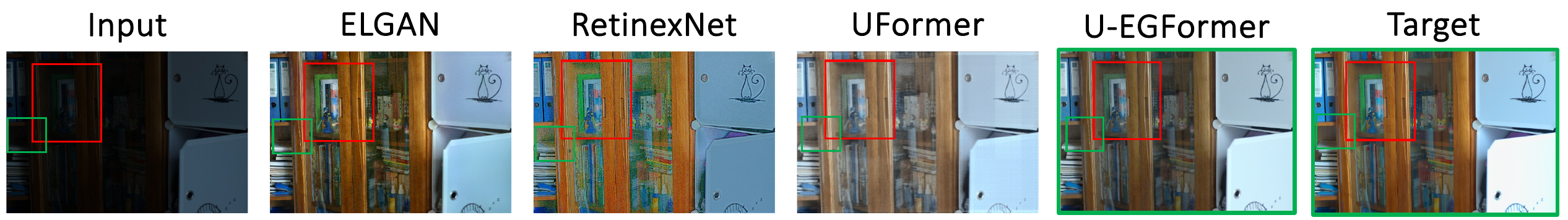} \vspace{-0.5em}
        \includegraphics[width=\linewidth]{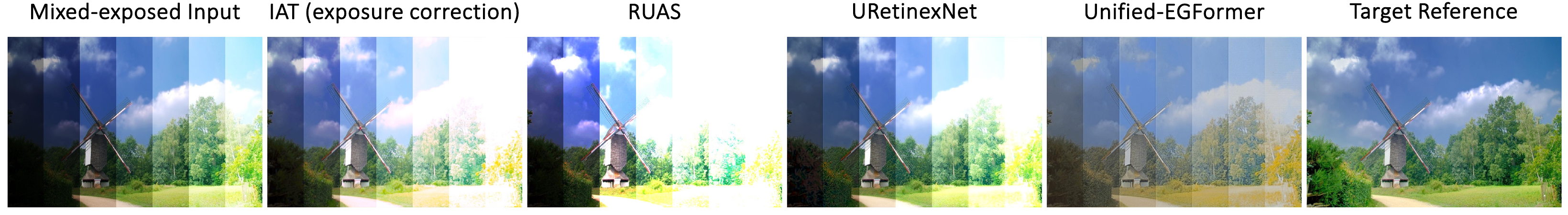}
        \caption{Qualitative comparison: \textbf{[Top]} Our method with competitive baselines on the low-light image enhancement task (image size 400x600). \textcolor{green}{green}:~comparing noise; \textcolor{red}{red}:~comparing color. \textbf{[Bottom]} Our method with other competitive baselines on mixed-exposure synthetic gradient dataset  (image size 900x600.) 
        }
        \label{fig:lolv1-sice-grad-results}
    \end{figure*}
    \begin{wrapfigure}[10]{L}{0.5\textwidth}
        \vspace{-2em}
        \centering
        \includegraphics[width=0.93\textwidth]{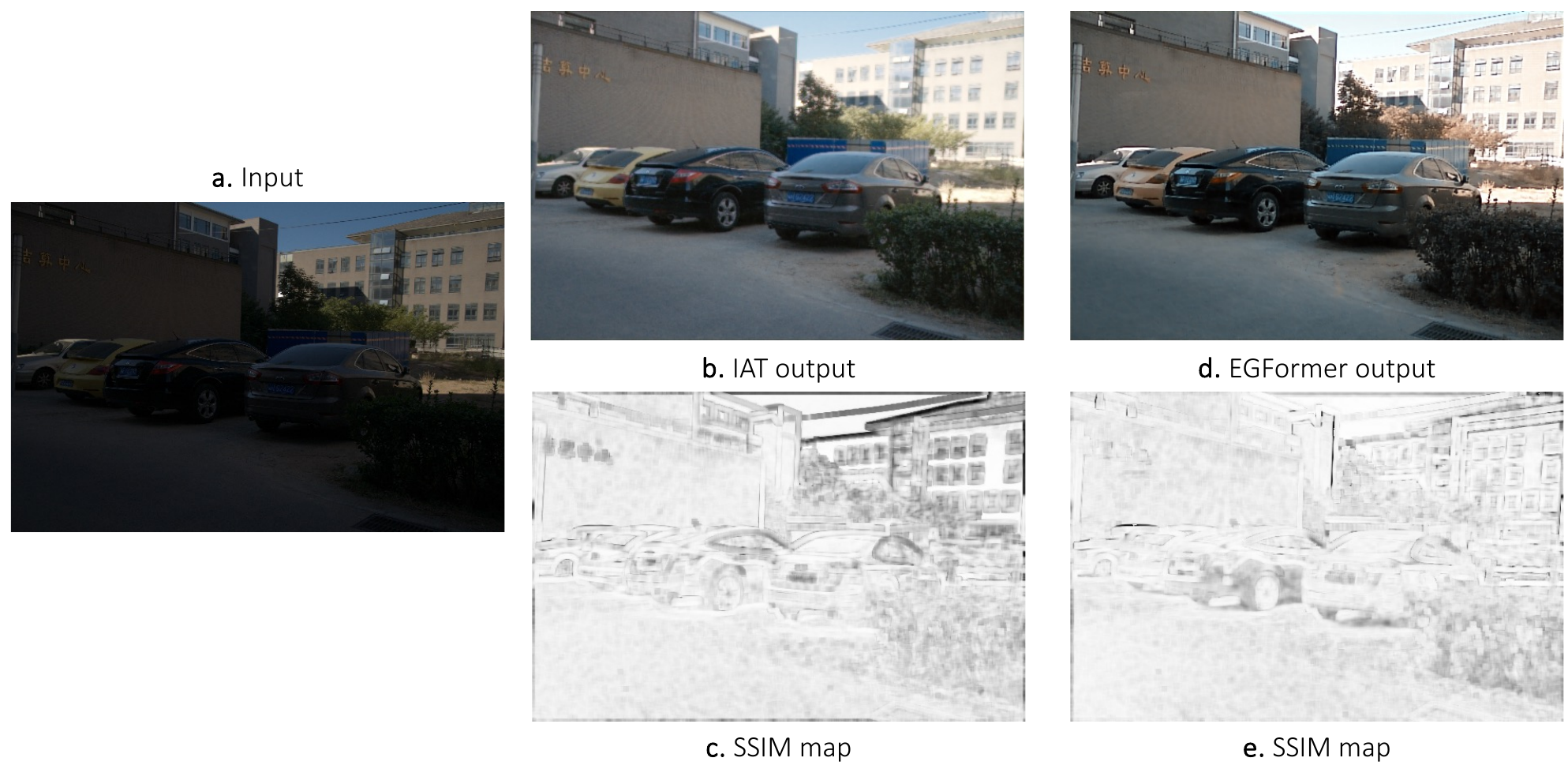}
        \caption{A comparison of the SSIM-maps for IAT \cite{Cui_2022_BMVC} and the proposed model.}
        \label{fig:ssim-map-comparison}
    \end{wrapfigure}
    \begin{table}[ht]
\centering
\caption{Visual results of our module settings in the pipeline over LOLv2-real dataset. ``\xmark''(resp.``\cmark'') means the module was unused (resp. used). `*' represents multiple warmup restarts. 
\small{\cgmark: $L_{1} + \text{MSE} + KL_{\mathcal{P}} + \text{SSIM}$,~
\cbmark: $L_{1} + \text{MSE} + \text{KL}_{\mathcal{P}} + \text{SSIM} + \text{VGG}$,~
\cymark: $L_{1} + \text{MSE} + \text{KL}_{\mathcal{P}} + \text{SSIM} + \text{Mul-Add}$,~
\cmark:  $L_{1} + \text{MSE} + \text{KL}_{\mathcal{P}} + \text{SSIM} + \text{VGG} + \text{MA-SL}_{1}$ }}
\label{tab:ablation-study}
\scalebox{0.85}{
\begin{tabular}{|l|cccccccc|}
    \hline
    \multirow{2}{*}{\textbf{Components}} & \multicolumn{8}{c|}{\textbf{Settings}} \\ \cline{2-9}
        & \textbf{1} & \textbf{2} & \textbf{3} & \textbf{4} & \textbf{5} & \textbf{6} & \textbf{7} & \textbf{8} \\ \hline

    Warmup
        & \xmark & \xmark & \cmark & \cmark & \cmark & \cmark & \cmark & \cmark* \\
    
    LEB (Attention map)
        & \xmark & \cmark & \cmark & \cmark & \cmark & \cmark & \cmark & \cmark \\

    GEB (Attention map)
        & \xmark & \xmark & \cmark & \xmark & \xmark & \xmark & \xmark & \xmark \\

    GEB (Inverse Illuminance map + input)
        & \xmark & \xmark & \xmark & \xmark & \xmark & \xmark & \xmark & \cmark \\
    
    GEB (Illuminance map + input)
        & \xmark & \xmark & \xmark & \cmark & \cmark & \cmark & \cmark & \xmark \\

    Attention Transformer Block
        & \xmark & 3 & 8 & 8 & 5 & 5 & 5 & 5 \\

    EAF Block
        & \xmark & \xmark & \xmark & \xmark & \xmark & \xmark & \xmark & \cmark \\

    Mul-Add Loss
        & \xmark & \cgmark & \cgmark & \cbmark & \cbmark & \cymark & \cmark & \cmark \\
    
    Adaptive Gamma net.
        & \cmark & \xmark & \xmark & \cmark & \cmark & \cmark & \cmark & \cmark \\ \hline
    
    \rowcolor{gray!20}\textbf{PSNR $\uparrow$}
        & 21.12 & 18.28 & 18.82 & 20.32 & 20.66 & 21.33 & 21.90 & 22.05 \\

    \rowcolor{gray!20}\textbf{SSIM $\uparrow$}
        & 81.56 & 72.12 & 74.24 & 81.96 & 81.36 & 79.86 & 83.92 & 84.10 \\ \hline
\end{tabular}}
\end{table}

    The LOL dataset is a challenging dataset even for state-of-the-art models due to its extremely low-light scenario. In Fig.~\ref{fig:lolv1-sice-grad-results} \textit{top}, we compare recent top models, where most of the models fail to match the color of the wood pane in this case with a lower PSNR score. In Fig.~\ref{fig:lolv1-sice-grad-results} \textit{bottom}, we show visual results for the SICE Grad dataset for the mixed-exposure task. In Supplementary Material (Fig.~12)
    we show a few challenging examples where LEB and GEB alone could not manage certain cases with extreme low and bright pixels, where the EAF block helps in re-highlighting the attended features. 

    \vspace{8pt}
    \noindent \textbf{Quantitative Comparison.}~ Tab.~\ref{tab:mev2_sicev2} reports PSNR and SSIM scores for U-EGformer and $\text{U-EGformer}_{eaf}$. U-EGformer demonstrates superior performance in handling both underexposure and overexposure scenarios across ME-v2 and SICE-v2 datasets, outperforming the majority of existing methods with significantly fewer parameters. Despite SID-L \cite{huang2022exposure} utilizing $>10M$ parameters we perform quite similarly (with differences of 0.0398/0.05 for SSIM/PSNR), while U-EGformer is \textbf{115 times} smaller network than SID-L. In Tab.~\ref{tab:lolv1_lolv2_FiveK}, we show remarkable generalization across LOL-v1, LOL-v2, and MIT-FiveK datasets, outperforming many baselines and illustrating robustness in exposure correction. Moreover, Tab.~\ref{tab:sicegrad_sice_mix} sets new benchmarks on the challenging SICE Grad and SICE Mix datasets, underscoring U-EGformer's superior performance in correcting mixed exposure images. In Fig.~\ref{fig:ssim-map-comparison}, we show a direct comparison of SSIM maps over the enhanced outputs of  IAT and U-EGformer. Darker pixels in SSIM maps as seen more in IAT than in U-EGformer, indicate areas where the enhanced outputs from the two frameworks significantly differ with ground-truth. Moreover, in Fig.~\ref{fig:quant-comparison-lol}, we demonstrate the noise and the color consistency that visually seems better in U-EGformer's output.

    \vspace{8pt}
    \noindent \textbf{Adaptable Learning Across Diverse Exposures.~}\label{sec:unified-framework} 
    Tackling the challenge of dataset diversity, our methodology enhances transferability and adaptability of learned models. Leveraging mechanisms such as attention masks allows us to consider simultaneously varying exposure conditions. Our unified framework demonstrates enhanced generalization capabilities, enabling effective fine-tuning across different datasets. Evidence of this robust adaptability is showcased in Tab.~\ref{tab:sicegrad_sice_mix}, illustrating our model's consistent performance on varied datasets with minimal fine-tuning adjustments.

    \vspace{8pt}
    \noindent \textbf{Ablation Study.}~ Our framework utilizes a data-centric approach with a smaller memory footprint ($\sim$12.5 Mb\footnote{over LOL-v2 input image}) and computation alongside other strategies as we have shown through Tab.~\ref{tab:mev2_sicev2}'s `\#params' column. Through Tab.~\ref{tab:ablation-study}, we show the effectiveness of each module in our framework over LOL-v2 dataset. We demonstrate that the inverse illuminance map, combined with the attention map and exposure-aware fusion block, achieves the best results when configured with the appropriate combination of loss functions. The first column achieves better performance on LOLv2.

\section{Conclusion}
    Our work introduces Unified-EGformer, addressing mixed exposure challenges in images with a novel transformer model. Through specialized local and global refinement alongside guided attention, it demonstrates superior performance across various scenarios. Its lightweight architecture makes it suitable for real-time applications, advancing the field of image enhancement and restoration.  Unified-EGformer could be enhanced further by refining the attention mechanism to become color independent to diminish the influence of color artifacts in the enhanced output. Additionally, exploring the integration of lightweight state space models \cite{gu2024mamba,adhikarla2024expomamba}, with bi-exposure guidance offers promising avenues for further optimizing the network for efficiency and performance in image enhancement tasks.

%
%
%
\bibliographystyle{splncs04}
\bibliography{references}

\clearpage
\setcounter{page}{1}
\section*{Supplementary Material}
\label{ref:Supp}

\section{Discussion \& Future Work}
    
    The challenges presented by mixed exposure in images are not just specific to any single methodology but are a broader issue within the field of image enhancement. As illustrated in Fig.~\ref{fig:attentionmap-thresholding}, both overexposure and underexposure present unique challenges. Overexposure often results in the loss of critical details such as edges and contours, while underexposure can hide essential information in darkness. These scenarios make effective image reconstruction a complex task.
    
    A further complication arises when considering the standard for evaluation—what exactly constitutes an ideal ground truth in the context of mixed exposure? Often, ground truth images themselves may lack details in over-exposed areas, complicating the assessment of enhancement algorithms as can be seen in Fig.~\ref{fig:attentionmap-thresholding}. This highlights a significant gap in our current understanding and capabilities, emphasizing the need for advancements that can precisely discern and correct varying degrees of exposure while preserving the integrity of the image details.
    
    The field of mixed exposure is not new; however it has seen limited exploration and is thus far from successfully solving the problem. Future research could consider integrating Kolmogorov-Arnold Networks (KANs) \cite{liu2024kan} in place of MLP blocks in transformers as they can yield more efficient and explainable models. KANs’ adaptive activation functions can better distinguish and process overexposed and underexposed regions, enhancing image quality.
    
    While current methods address many aspects of image quality, there is still room for improvement in creating color-independent ground truths for guided attention map generators. Color-independent techniques prevent issues such as greens appearing yellow due to color influence, ensuring accurate hue representation and processing. Understanding and correcting these color dynamics may require 
    using attention map generators to identify inconsistencies.
    
    Such methods must not only navigate the complexities introduced by mixed exposure but also contribute to a deeper understanding of what ideal image enhancement should entail in diverse real-world conditions.

\section{Model Design Components}
    This extended material provides more details of our model design components.

\subsection{Algorithm}
    The details presented in the training are shown in Algorithm \ref{alg:egformer} below:
    \begin{algorithm}[htb]
        \SetCommentSty{mycommfont}
        \SetKwInOut{Input}{Input}
        \SetKwInOut{Output}{output}
        \Input{A collection of mixed-exposure images $\textbf{D}$, Guided attention map generator $\mathbf{M}_g$, illumination map generator $\mathbf{I}_g$, local enhancement block \textbf{LEB}, global enhancement block \textbf{GEB}, exposure-aware attention fusion) block \textbf{EAF}, training epochs $E$.}
        \BlankLine
        \nonl \textbf{Manual pixel-level Otsu thresholding mask:} \\
        Initialize binary class mean with $\mu$ \\
        \ForPar{\textup{image} $d \in \mathbf{D}$}
        {\textup{Calculate $\mu_d$ via Otsu's method}}
        $\mu = \sum_{d \in \mathbf{D}} \mu_d / |\mathbf{D}|$ \\
        Apply $\mu$ manually to $d \in \mathbf{D}$ to get binary masks $\mathbf{D}_m$ for identifying exposure. \\
        \BlankLine
        \nonl \textbf{Illumination map inversion:}\\
        $\mathbf{D}_i \leftarrow$ Operation($\mathbf{D}$)\\
        \BlankLine
        \nonl \textbf{EGformer executes} \var{Training$(\mathbf{D}, \mathbf{M}_g, \mathbf{I}_g, \mathbf{W}_{LEB}, \mathbf{W}_{GEB}, \mathbf{W}_{EAF}, E)$}\textbf{:}\\
        \For{\textup{each epoch} $e = 1,2,...,E$}{
        \For{\textup{each batch} $\mathbf{b} \in \mathbf{D}$, $\mathbf{b}_m \in \mathbf{D}_m$, $\mathbf{b}_i \in \mathbf{D}_i$}{
            $\nabla \mathcal{L}_m \leftarrow \textit{Attention-Loss}(\mathbf{M}_g(\mathbf{b}), \mathbf{b}_m)$  \\

            $\nabla \mathcal{L}_i \leftarrow \textit{Pretraining loss Eq.~\ref{eq:pretraining}~}(\mathbf{I}_g(\mathbf{b}), \mathbf{b}_i)$  \\

            $\mathcal{L} \leftarrow \mathcal{L}_m + \mathcal{L}_i$ \\
            $\text{update} ~\theta$
        }}
        \For{\textup{each epoch} $e = 1,2,...,\hat{E}$}{
        \For{\textup{each batch} $\mathbf{b} \in \mathbf{D}$, $\mathbf{b}_m \in \mathbf{D}_m$, $\mathbf{b}_i \in \mathbf{D}_i$}{
            
            $\nabla \mathcal{L}_m \leftarrow \textit{Attention-Loss}(\mathbf{M}_g(\mathbf{b}), \mathbf{b}_m)$  \\

            $\nabla \mathcal{L}_i \leftarrow \textit{finetuning-loss Eq.~\ref{eq:finetuning}~}(\mathbf{I}_g(\mathbf{b}), \mathbf{b}_i)$  \\

            $\mathcal{L} \leftarrow \mathcal{L}_m + \mathcal{L}_i$ \\

            $\text{update} ~\theta$
        }}
        return $~\theta$
        \caption{U-EGformer Framework for Image Enhancement}
        \label{alg:egformer}
    \end{algorithm}

\subsection{A-MSA} 
\label{sm-sec:a-msa}

    The A-MSA component can be mathematically represented as:
    \begin{equation}
    \label{mapgen-eq}
        \begin{split}
            \mathcal{A}(x) &= \text{Softmax}\left(\frac{\mathcal{Q}_n(m)\mathcal{K}_n(m)^T}{\sqrt{d_k}}\right)\mathcal{V}_n(m), \\
            & n \in \{1, 2, \ldots, N\}
        \end{split}
    \end{equation}
    where $\mathcal{A}(m)$ is the attention map, $\mathcal{Q}(m)$, $\mathcal{K}(m)$, and $\mathcal{V}(m)$ denote the query, key, and value projections of the input $m$, and $d_k$ is the dimensionality of the key, thereby guiding the model to focus on relevant features, when and how much attention should be paid to each element in the sequence when processing a query.

\subsection{Dual Gating Feedforward Network (DGFN).~} 
\label{sm-sec:dgfn}

    The DGFN mechanism involves two parallel pathways that process the input features $m \in \mathbf{M}_g$ through distinct gating mechanisms, utilizing GELU activations ($\phi$) and element-wise products. Each path applies a sequence of convolutional transformations to enrich local context, comprising a \(1 \times 1\) convolution followed by a \(3 \times 3\) depth-wise convolution. Moreover, depth-wise convolutions facilitate spatial information processing while significantly reducing computational costs, making them particularly suitable for developing a light-weight environment without compromising performance. Following the approach in Wang et al.~\cite{wang2023ultra}, the outputs of these parallel pathways are then merged using element-wise summation, allowing for a comprehensive feature refinement that incorporates both detailed and global information.

    Mathematically, the operation of DGFN on input features \(Y\) can be formulated as follows:
    \[
        \hat{M} = \phi\left(W_{1 \times 1}^{(1)} M\right) \odot \phi\left(W_{3 \times 3}^{(1)} W_{1 \times 1}^{(1)} M\right) + \phi\left(W_{1 \times 1}^{(2)} M\right) \odot \phi\left(W_{3 \times 3}^{(2)} W_{1 \times 1}^{(2)} M\right),
        \label{eq:dgfn}
    \]
    where \(W_{1 \times 1}^{(1)}, W_{3 \times 3}^{(1)}, W_{1 \times 1}^{(2)},\) and \(W_{3 \times 3}^{(2)}\) denote the convolutional filters in the two pathways, and \(\odot\) represents element-wise multiplication. The final output \(\hat{M}\) thus combines the processed features from both pathways, ensuring that the network selectively emphasizes informative regions for enhanced exposure correction in the attention map generation process.
    
    This design allows the network to make more nuanced adjustments to the attention maps, focusing on areas of the image that require exposure correction while maintaining the integrity of well-exposed regions.

\subsection{Model Training}
\label{sm-sec:dataset-specific-training}

    For LOL-v1 \cite{WeiWY018}, we adopt the standard split provided, comprising images of dimensions $400 \times 600$, with 500 images allocated for both training and evaluation. Similarly, for LOL-v2 \cite{WeiWY018}, we utilize the real-world dataset for both training and evaluation, consisting of images of size $400 \times 600$, with 689 and 100 images in the respective splits. We adopt patch sizes of $256 \times 256$ and $324 \times 324$ for training across both LOL-v1 and LOL-v2 datasets. Additionally, we incorporate the MIT-Adobe FiveK \cite{fivek} dataset for comparative analysis with other baseline methods. The FiveK dataset comprises 5,000 image pairs, each meticulously retouched by experts (A - E), with the best-performing result reported. To align with established practices \cite{Cui_2022_BMVC,afifi2021learning}, we crop the training images to $256 \times 256$ dimensions, subsequently fine-tuning over $324 \times 324$ and $512 \times 512$ patches, while the test images are resized to a maximum dimension of 512 pixels. Additionally, we incorporate the MSEC \cite{afifi2021learning} dataset, which features realistic underexposure and overexposure scenarios derived from the MIT-FiveK dataset. Our experimental setup involves 17,675 training images, 750 for validation, and 5,905 for testing, adhering to the general experimental protocol of our baselines. Furthermore, we utilize the SICE \cite{Cai2018deep} dataset, employing the SICE-v2 variant, a resized version of the original dataset, following the approach outlined in \cite{huang2022exposure-consistency,huang2023learning}. Lastly, to comprehensively evaluate mixed-exposure scenarios, we assess our model's performance on two variants of the SICE dataset: SICE-Grad and SICE-Mix \cite{zheng2022low}, which feature mixed under and overexposure regions within single images.

    To facilitate network training, we employ the cosine annealing schedule along with the Cosine-Annealing learning rate scheduler, incorporating 15 warm-up epochs and setting eta\_min to $1e-5$. Additionally, we utilize the Adam optimizer with $\beta_1 = 0.9$ and $\beta_2 = 0.999$, along with an epsilon value of $1e-8$. Our initial learning rate is set to $1e-4$ for pretraining purposes. Furthermore, we adopt a batch size of 4 and a weight decay of $1e-4$ to stabilize training and improve model performance. 

\section{Extended Experimental Results}
\subsection{Generalization in the Wild}
\label{sec:ab_wild}

    Our model demonstrates impressive visual qualitative results on the BAID dataset. Initially trained on the LOLv2 dataset, the model was finetuned using only 300 randomly selected images from the BAID train dataset, out of the 3000 available backlit images. These BAID images were originally downsampled images and were directly used from Liang et al.~\cite{liang2023iterative}. The backlit images in BAID present a challenging scenario due to uneven exposure, typically featuring underexposure in the foreground and overexposure in the background. Traditional approaches often struggle with these conditions, but Unified-EGformer, particularly with our Global Enhancement Block (GEB) combined with the Exposure Aware Fusion Block (EAF), excel by exploiting contextual features. Despite the limited finetuning data, our model significantly enhances backlit images, effectively generalizing to both indoor and outdoor scenes under diverse backlit conditions.

    \begin{figure}[htb]
        \centering
        \includegraphics[angle=90, width=0.95\textwidth]{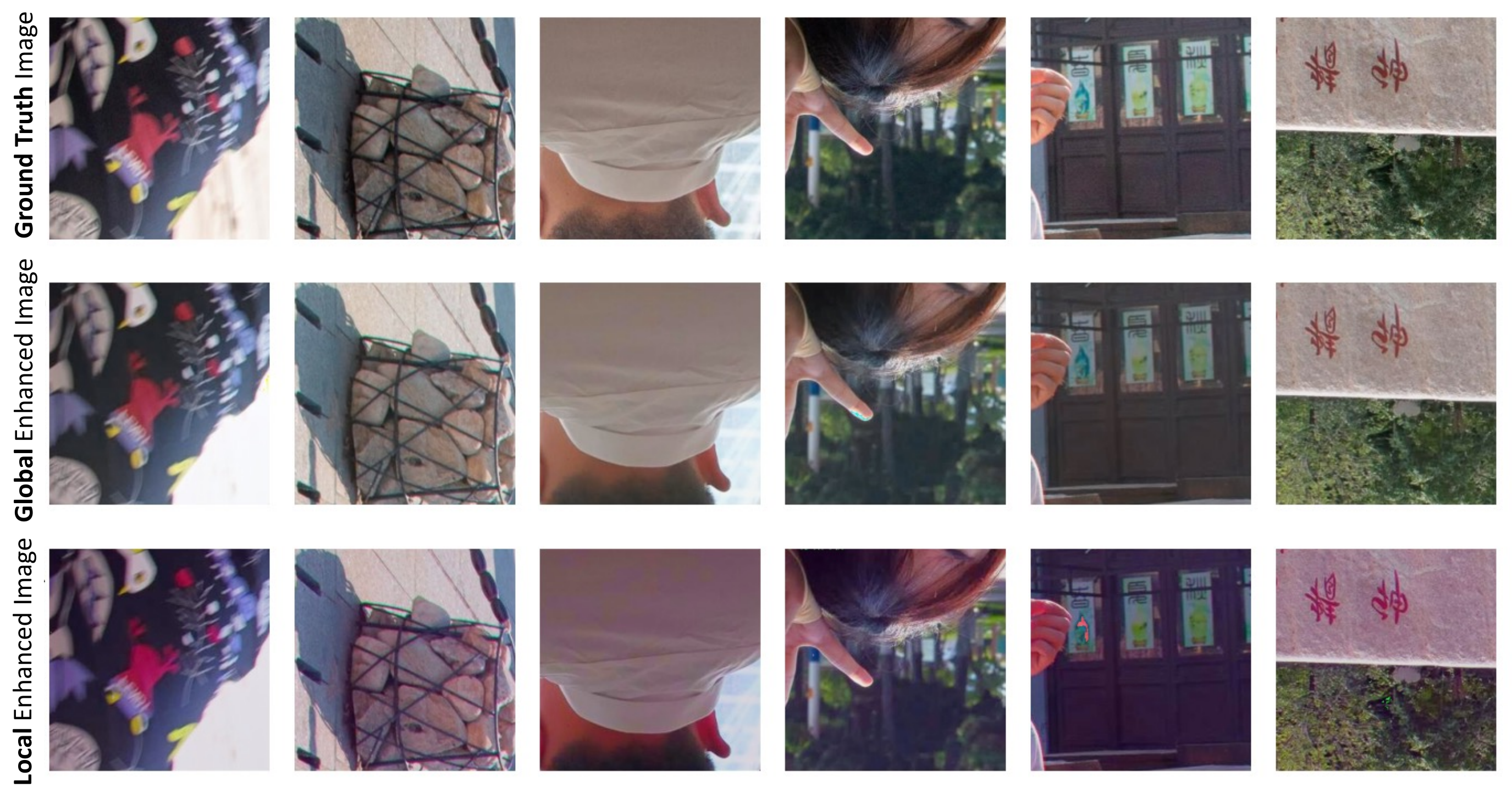}
        \caption{U-EGformer's output from `Local Enhancement Block (LEB)' and `Global Enhancement Block (GEB) + Exposure Aware Fusion Block (EAF)' enhancement modules compared with Ground-truth image on BAID set.}
        \label{fig:baid-test-generalization}
    \end{figure}
    \begin{figure}[htb]
        \centering
        \includegraphics[width=\textwidth]{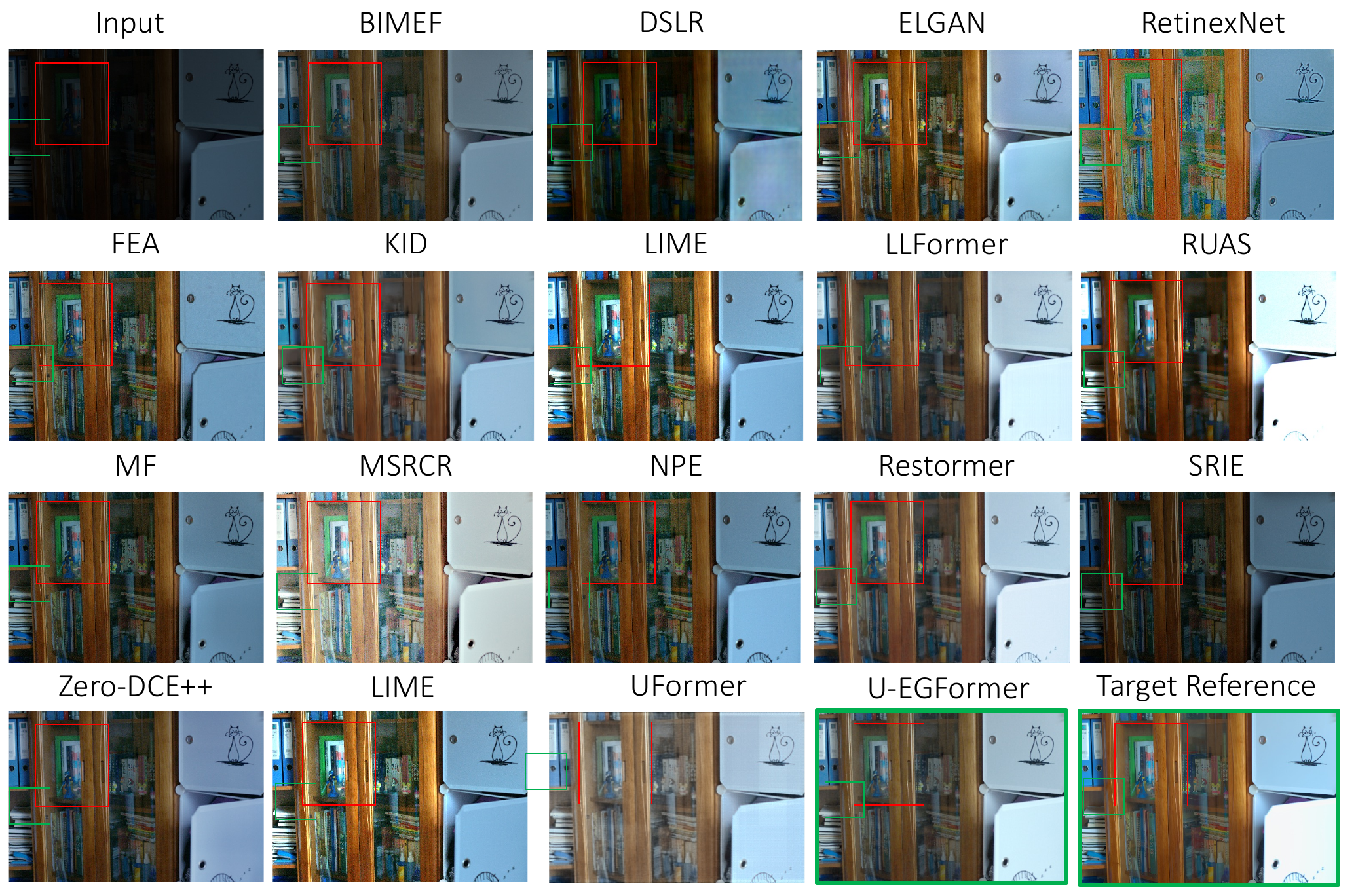}
        {\caption{\label{fig:quant-comparison-lol}Qualitative comparison of our proposed method with competitive baselines on low-light image enhancement task over the image size 400x600. \textcolor{green}{green}:~comparing noise; \textcolor{red}{red}:~comparing color.}}
        \label{fig:lolv1-output-comparison-full}
    \end{figure}
    \begin{figure}[htb]
        \centering
        \includegraphics[width=\linewidth]{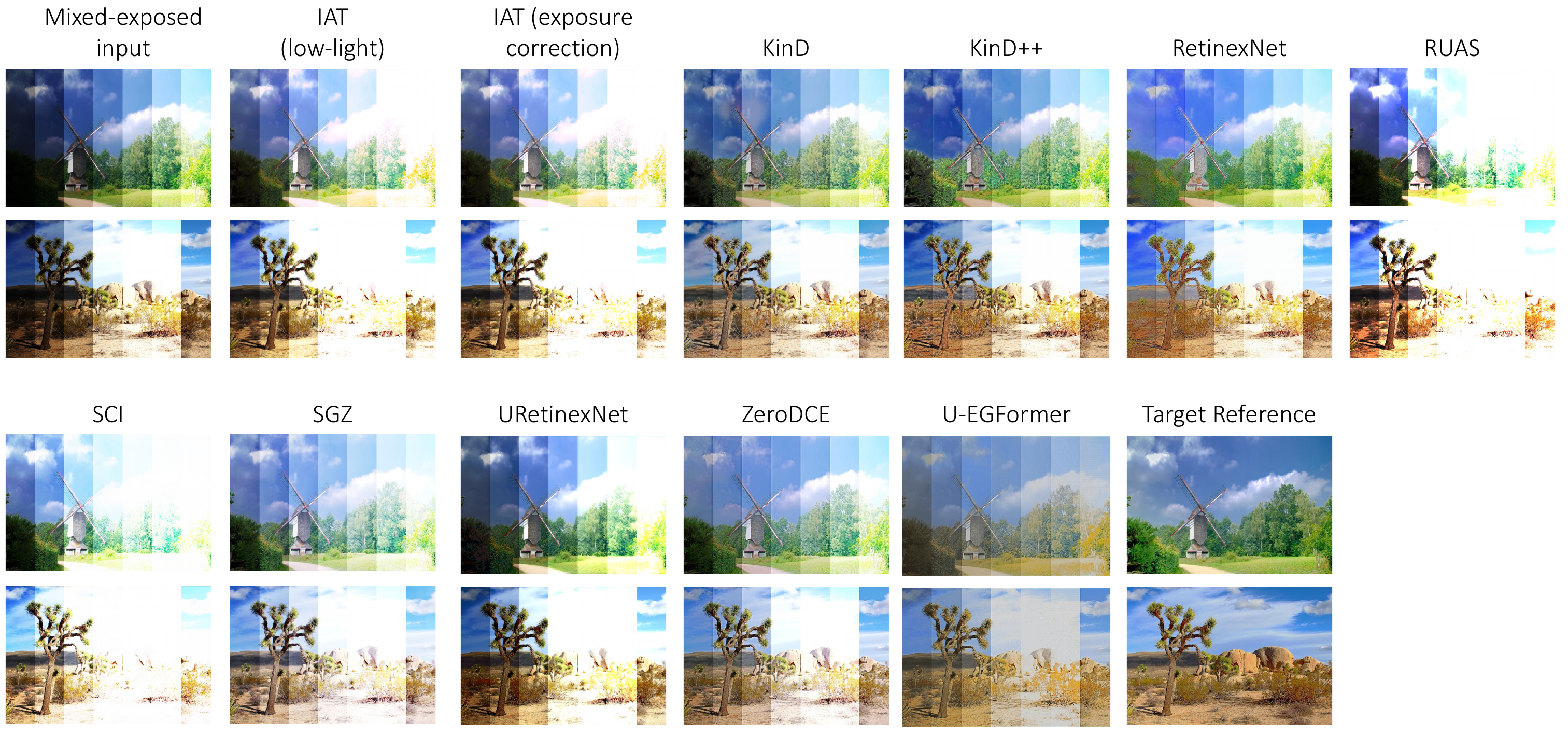}
        \caption{Qualitative comparison of our proposed method with other competitive baselines on mixed-exposure synthetic gradient dataset for image enhancement task over the image size 900x600.}
        \label{fig:sicegrad-output-comparison-full}
    \end{figure}
    \begin{figure}[htb]
        \centering
        \includegraphics[width=\textwidth]{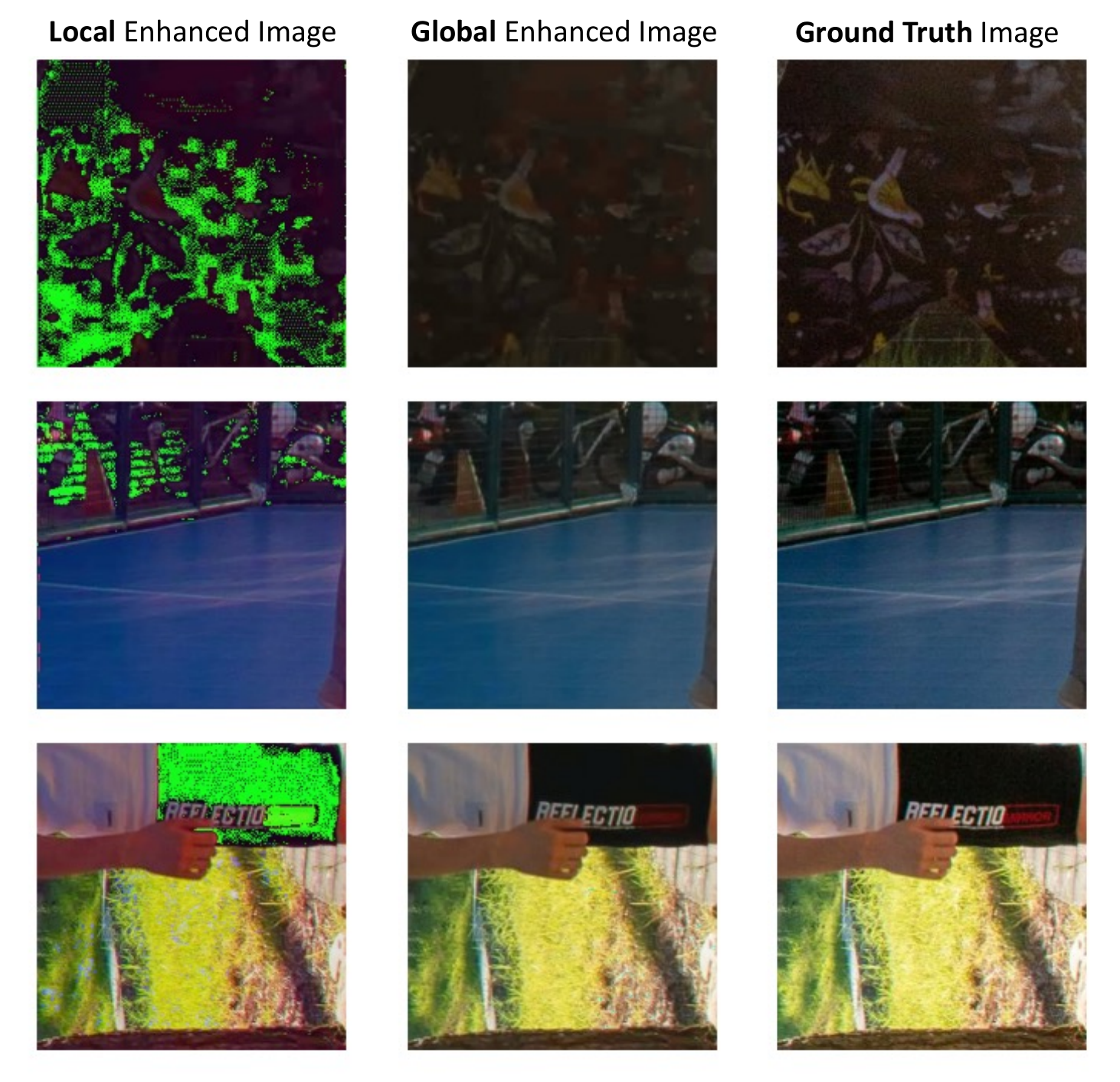}
        \caption{Showing the Global+EAF block's significance when adverse lighting brings artifacts in the reconstructed image. First column: `Local Enhancement Block (LEB)' and second column: `Global Enhancement Block (GEB) + Exposure Aware Fusion Block (EAF)' enhancement modules compared with third column: Ground-truth image on BAID test set.}
        \label{fig:fusion-block}
    \end{figure}

\end{document}